\pdfoutput=1

\documentclass[11pt]{article}

\usepackage[preprint]{acl}

\usepackage{times}
\usepackage{longtable}
\usepackage{latexsym}
\usepackage{booktabs}
\usepackage{adjustbox}
\usepackage{multicol}
\usepackage{multirow}
\usepackage{amsmath}
\usepackage{amsfonts}
\usepackage{amssymb}
\usepackage{pdflscape}
\usepackage{philex}
\usepackage{enumitem}
\usepackage{bbm}

\usepackage{hhline}
\usepackage[T1]{fontenc}

\usepackage[utf8]{inputenc}

\usepackage{microtype}

\usepackage{tcolorbox}

\usepackage{inconsolata}

\usepackage{graphicx}

\usepackage{subcaption}

\title{Adaptive Retrieval without Self-Knowledge?\\ Bringing Uncertainty Back Home}

\author{
 \textbf{Viktor Moskvoretskii\textsuperscript{1,3}},
 \textbf{Maria Lysyuk\textsuperscript{2,1}},
 \textbf{Mikhail Salnikov\textsuperscript{2,1}},
 \textbf{Nikolay Ivanov\textsuperscript{1}},
\\
 \textbf{Sergey Pletenev\textsuperscript{2,1}},
 \textbf{Daria Galimzianova\textsuperscript{4}},
 \textbf{Nikita Krayko\textsuperscript{4}},
 \textbf{Vasily Konovalov\textsuperscript{2,5}},
\\
 \textbf{Irina Nikishina\textsuperscript{6}},
 \textbf{Alexander Panchenko\textsuperscript{2,1}}
\\
 \textsuperscript{1}Skoltech,
 \textsuperscript{2}AIRI,
 \textsuperscript{3}HSE University,
 \textsuperscript{4}MTS AI,
 \textsuperscript{5}MIPT,
 \textsuperscript{6}University of Hamburg
\\
 \small{
   \textbf{Correspondence:} \href{mailto:vvmoskvoretskii@gmail.com}{vvmoskvoretskii@gmail.com}
 }
}

\begin{document}
\maketitle
\begin{abstract}
Retrieval Augmented Generation~(RAG) improves correctness of Question Answering (QA) and addresses hallucinations in Large Language Models~(LLMs), yet greatly increase computational costs. Besides, RAG is not always needed as may introduce irrelevant information.  Recent adaptive retrieval methods integrate LLMs' intrinsic knowledge with external information appealing to LLM self-knowledge, but they often neglect efficiency evaluations and comparisons with uncertainty estimation techniques. 
We bridge this gap by conducting a comprehensive analysis of 35 adaptive retrieval methods, including 8 recent approaches and 27 uncertainty estimation techniques, across 6 datasets using 10 metrics for QA performance, self-knowledge, and efficiency. Our findings show that uncertainty estimation techniques often outperform complex pipelines in terms of efficiency and self-knowledge, while maintaining comparable QA performance.


\end{abstract}

\section{Introduction}

Large Language Models have gained increased popularity due to their remarkable performance across diverse tasks, such as question answering (QA)~\cite{DBLP:conf/emnlp/Yang0ZBCSM18, DBLP:journals/tacl/KwiatkowskiPRCP19}. At the same time, hallucinations represent a substantial challenge for LLMs. Solely utilizing only parametric knowledge in generating trustworthy content is limited by the knowledge boundaries of LLMs~\cite{DBLP:conf/acl/YinZR024}, which may potentially lead to internal hallucinations~\cite{DBLP:journals/corr/abs-2402-10612-rowen}. While external information via RAG \cite{DBLP:conf/nips/LewisPPPKGKLYR020} can potentially help to fill these gaps, it raises the possibility of irrelevance, thus leading to the error accumulation~\cite{DBLP:conf/icml/ShiCMSDCSZ23} and increasing the likelihood of external hallucinations~\cite{DBLP:journals/corr/abs-2402-10612-rowen}. 

To balance between the intrinsic knowledge of LLMs and external information, adaptive retrieval methods have emerged~\cite{DBLP:conf/acl/SuTA0024, DBLP:journals/corr/abs-2402-10612-rowen, DBLP:conf/naacl/JeongBCHP24}. These methods rely on LLM \textbf{self-knowledge} --- model capacity to recognize its own knowledge~\cite{yin2023large} --- and determine when it lacks critical information.

\begin{figure}[t!]
    \centering
    \includegraphics[width=\linewidth]{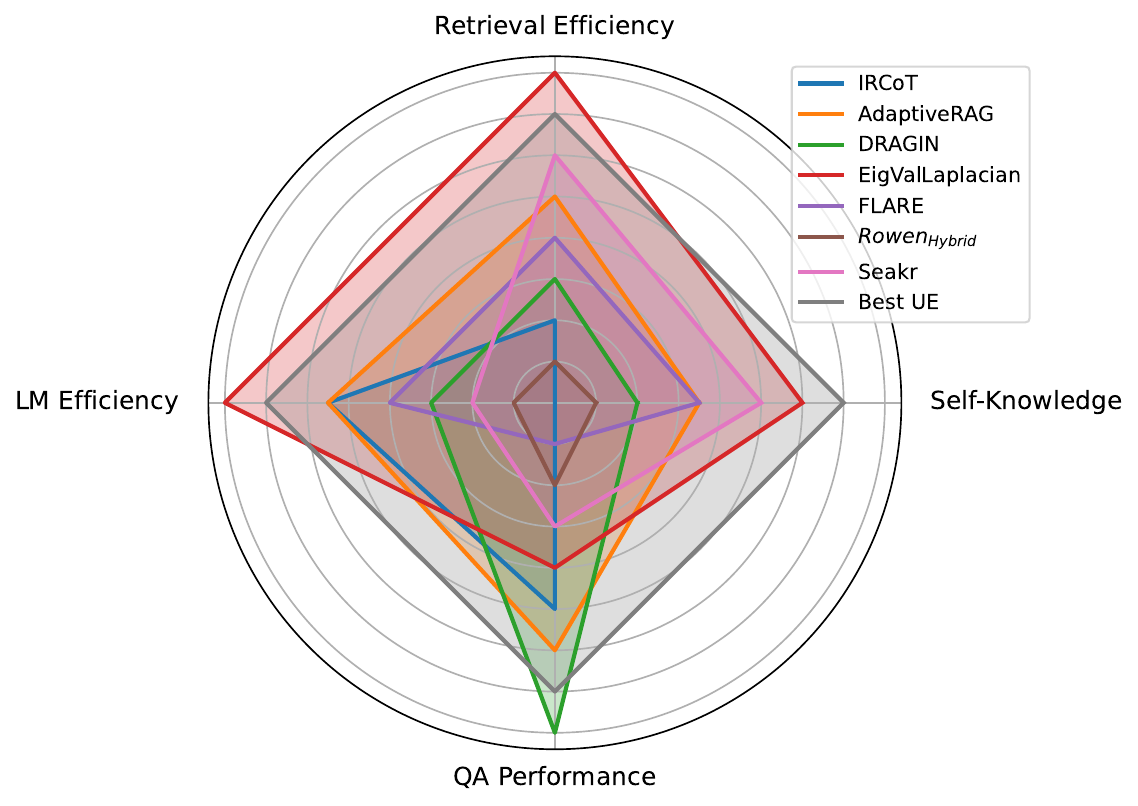}
    \caption{Performance comparison of the state-of-the art models across efficiency metrics (number of LLM calls, Retrieval calls), QA quality metric (In-Accuracy), and the ability to identify self-knowledge (ROCAUC). The plot demonstrates the reverted ranks of the methods across 6 datasets.}
    \label{fig:radar}
\end{figure}

Adaptive retrieval methods may not only improve answer correctness, but also substantially decrease retrieval calls, enhancing efficiency. While recent methods have focused extensively on the retrieval calls~\cite{DBLP:conf/acl/SuTA0024, DBLP:conf/naacl/JeongBCHP24, DBLP:conf/acl/TrivediBKS23}, they often overlook the cost of LLM calls, which can be even more expensive, especially with proprietary models. Furthermore, recent studies of complex pipelines do not assess self-knowledge abilities and lack comparisons with well-established uncertainty estimation methods, such as  Mean Entropy~\cite{fomicheva2020unsupervised}.

To address these limitations, 
we conduct a comprehensive study of 35 adaptive retrieval systems, including 8 recently published methods and 27 established uncertainty estimation methods, across 6 QA datasets covering both simple one-hop and complex multi-hop questions. We evaluate these methods in terms of 
the QA performance, self-knowledge, and two types of efficiency, using a total of 10 metrics. Our evaluation, shown in Figure~\ref{fig:radar}, reveals that no single method dominates across all axes. However, well-established uncertainty estimation methods are often more useful compared to recently published, more complex pipelines.

Finally, we provide a rigorous in-depth assessment of the out-of-distribution (OOD) performance of uncertainty methods and analyze the complexity of their functional classes.

Our contributions and findings are as follows: 

\begin{enumerate}[itemsep=0.3pt,topsep=0.5pt]
    \item A consistent study of 35 adaptive retrieval methods on 6 single- and multi-hop datasets, evaluating QA performance, self-knowledge, and efficiency across 10  metrics.
    \item The first comprehensive application and comparison of 27 well-established uncertainty estimation methods for adaptive retrieval, showcasing their potential and efficiency.
    \item An in-depth analysis of uncertainty methods for adaptive retrieval, covering OOD transfer and examining the complexity of their functional classes.
\end{enumerate}

We publish all the code and data.\footnote{\url{https://github.com/s-nlp/AdaRAGUE}}

\section{Related Work}

\paragraph{Retrieval-Augmented Generation} methods are widely used to enhance the performance of LLMs in many tasks, like up-to-date information~\cite{jiang2024learningeditaligningllms} or questions about rare entities in which LLM shows poor generation quality due to lack of inner knowledge~\cite{allenzhu2024physicslanguagemodels31}.
In the simplest case, the input sequence of the question is used as a query for databases or search engines. The resulting information is then incorporated as an additional context, proven effective for a variety of tasks ~\cite{Khandelwal2020Generalization,NEURIPS2020_6b493230} and models~\cite{pmlr-v162-borgeaud22a,ram-etal-2023-context,socher-etal-2013-recursive}. All these methods are applied to the retrieval once before generation, so they are often combined under the name single-round retrieval augmentation.

\paragraph{Adaptive Retrieval-Augmented Generation} methods perform retrieval for every query may be both inefficient and unnecessary. Moreover, retrieving knowledge at every step may be misleading or even conflicting with LLM's parameters~\cite{simhi2024constructingbenchmarksinterventionscombating}. Adaptive retrieval methods have emerged as an attempt to understand whether LLM needs external knowledge by exploiting models' self-knowledge abilities.

The decision to retrieve may depend on different criteria. It may be based on the text outputs of LLMs~\cite{DBLP:conf/acl/TrivediBKS23} or text consistency~\cite{DBLP:journals/corr/abs-2402-10612-rowen}, on the self-aware uncertainty of LLMs from their internal states~\cite{jiang-etal-2023-active, DBLP:conf/acl/SuTA0024, DBLP:journals/corr/abs-2406-19215} or using a trained classifier to decide whether to retrieve~\cite{DBLP:conf/naacl/JeongBCHP24}.




\paragraph{Uncertainty Estimation (UE)} measures the confidence in LLM predictions and can be classified into white-box and black-box methods. White-box methods require access to internal model details, such as logits or layer outputs, and are divided into information-based (using token or sequence probabilities from a single model), ensemble-based (leveraging probabilities from different model versions), and density-based (constructing a probability density from latent representations). Black-box methods, in contrast, only require access to the model’s output~\cite{DBLP:conf/emnlp/FadeevaVTVPFVGP23}.


\section{Methods}

In this section, we briefly introduce the existing adaptive retrieval methods. More details can be found in Appendix~\ref{sec:baselines_description}. 

\subsection{End-to-End Methods}

\textbf{IRCoT} (\textbf{I}nterleaving \textbf{R}etrieval in a \textbf{CoT}) is a dynamic approach that adds extra relevant passages from the retriever to the context if the current CoT step has not produced the answer yet. The query for extra context is based on the last generated CoT sentence~\cite{DBLP:conf/acl/TrivediBKS23}.

\textbf{Adaptive RAG}~\cite{DBLP:conf/naacl/JeongBCHP24} uses the classifier based on the T5-large model \cite{2020t5} that predicts one of the three outcomes: whether not to retrieve at all, retrieve once and retrieve multiple times with  IRCoT.

\textbf{FLARE} (\textbf{F}orward-\textbf{L}ooking \textbf{A}ctive \textbf{Re}trieval augmented generation) is a method that retrieves context when token probability falls below a threshold, regenerating the response until the next uncertain token or completion~\cite{jiang-etal-2023-active}.

\textbf{DRAGIN} (\textbf{D}ynamic \textbf{R}etrieval \textbf{A}ugmented \textbf{G}eneration based on \textbf{I}nformation \textbf{N}eeds) monitors token probabilities like FLARE but filters stopwords to identify uncertainty tokens. It improves context retrieval by reformulating queries using  attention weights and reasoning~\cite{DBLP:conf/acl/SuTA0024}.

\textbf{Rowen} (\textbf{R}etrieve \textbf{O}nly \textbf{W}hen It \textbf{N}eeds) is a consistency-based approach with two components: the Consistency Language, which measures answer consistency across English and Chinese, and the Consistency Model, which evaluates semantic coherence across models. Both output inconsistency scores to trigger retrieval. The Rowen Hybrid combines both components~\cite{DBLP:journals/corr/abs-2402-10612-rowen}.

\textbf{SeaKR} (\textbf{Se}lf-\textbf{a}ware \textbf{K}nowledge \textbf{R}etrieval) uses an Uncertainty Module (UM) to monitor LLM internal states and trigger retrieval when uncertainty exceeds a threshold. A re-ranking component selects a snippet that  reduces uncertainty and improves factual accuracy~\cite{DBLP:journals/corr/abs-2406-19215}.

\subsection{Uncertainty Estimation Methods}

For uncertainty estimation, we employ 27 different methods, described in detail in Table~\ref{tab:ue_desc}. In the main part of our paper, we focus on the 5 best-performing uncertainty estimation methods, which include approaches from various method families:

\begin{itemize}[itemsep=0.3pt,topsep=0.5pt]
    \item \textbf{Lexical Similarity}: Measures a consistency score based on the average similarity of sampled responses~\cite{fomicheva2020unsupervised}.
    \item \textbf{Max Entropy}: Computes the entropy of each token and aggregates it for the sequence using the maximum value~\cite{fomicheva2020unsupervised}.
    \item \textbf{Mean Entropy}: Computes the entropy of each token and aggregates it for the sequence using the mean value~\cite{fomicheva2020unsupervised}.
    \item \textbf{SAR}: Measures the entropy of each token, reweights it based on token relevance, and aggregates the values using a sum over the sequence~\cite{duan2023shifting}.

    \item \textbf{EigValLaplacian}: Computes the sum of Laplacian eigenvalues by constructing a weighted graph based on the consistency of sampled responses~\cite{lin2023generating}.
\end{itemize}

\section{Experimental Setup}

In this section, we briefly discuss the implementation details and the evaluation setup.

\subsection{Implementation Details}

We use the LLaMA 3.1-8b-instruct model~\cite{dubey2024llama} with the default generation parameters for all experiments. The baseline methods follow their original protocols, including prompting and parameter settings, while uncertainty estimation methods use the AdaptiveRAG protocol~\cite{DBLP:conf/naacl/JeongBCHP24}, with the same prompt and few-shot examples.

For all methods, we use the BM25 retriever~\cite{DBLP:conf/trec/RobertsonWJHG94} with Elasticsearch 7.17.9\footnote{\url{https://www.elastic.co/elasticsearch}} and the Wikipedia corpus preprocessed by~\citet{DBLP:conf/emnlp/KarpukhinOMLWEC20}, following previous studies~\cite{su2024dragindynamicretrievalaugmented,DBLP:journals/corr/abs-2406-19215}.

Uncertainty method scores are computed on both training and test sets using the LM-Polygraph~\cite{DBLP:conf/emnlp/FadeevaVTVPFVGP23}. A set of classifiers are trained on the training set scores, with the best classifier's performance reported based on downstream metrics. Additional details are provided in Appendix~\ref{sec:appendix_technical}.

\subsection{Datasets}

We use the single-hop and multi-hop QA datasets in the same experimental setup to replicate a real-world scenario where various queries have different difficulties. The choice of datasets is standard for the task with the single-hop questions -- SQUAD v1.1~\cite{DBLP:conf/emnlp/RajpurkarZLL16}, Natural Questions~\cite{DBLP:journals/tacl/KwiatkowskiPRCP19}, TriviaQA~\cite{DBLP:conf/acl/JoshiCWZ17}, and the datasets with the complex ones -- 
MuSiQue~\cite{DBLP:journals/tacl/TrivediBKS22}, HotpotQA~\cite{DBLP:conf/emnlp/Yang0ZBCSM18}, and 2WikiMultiHopQA~\cite{DBLP:conf/coling/HoNSA20}, following previous papers~\cite{DBLP:conf/acl/TrivediBKS23, DBLP:conf/naacl/JeongBCHP24, DBLP:conf/acl/SuTA0024, DBLP:journals/corr/abs-2406-19215}.
To ensure consistency, we use the subsets of 500 questions of the original test parts of the datasets from previous studies~\citep{DBLP:conf/acl/TrivediBKS23, DBLP:conf/naacl/JeongBCHP24}
.



\subsection{Evaluation}

We conduct a comprehensive evaluation using QA downstream metrics, efficiency metrics, and self-knowledge metrics to broadly cover every aspect of the model. To fairly compare performance across datasets, we also use methods ranks on each dataset (smaller rank indicates better performance) and average the ranks. This ensures a balanced evaluation, as performance gains may vary in significance across datasets.

\subsubsection{Downstream QA Metrics}

To assess the final QA system quality we use In-Accuracy, EM and F1, following previous studies~\cite{DBLP:conf/acl/MallenAZDKH23, DBLP:conf/emnlp/BaekJKPH23, DBLP:conf/iclr/AsaiWWSH24, DBLP:conf/naacl/JeongBCHP24}, where:

\begin{itemize}[itemsep=0.3pt,topsep=0.5pt]
    \item \textbf{In-Accuracy (InAcc)} evaluates whether the predicted answer includes the ground truth.
    \item \textbf{Exact Match (EM)} measures the exact match of prediction with the ground truth.
    \item \textbf{F1} quantifies the degree of token overlap between the predicted answer and the ground truth answer.
\end{itemize}

We primarily rely on In-Accuracy as the main metric, as it is more robust to answer variations compared to EM and provides a better measure of correctness than F1. Additionally, the overall trends across these metrics are generally consistent.

\subsubsection{Efficiency Metrics}

In addition to enhanced quality, adaptive retrieval procedures must also demonstrate improvements in efficiency; otherwise, consistent retrieval might remain superior. To evaluate it, we measure:

\begin{itemize}[itemsep=0.3pt,topsep=0.5pt]
    \item \textbf{Retriever Calls (RC)}: The average number of retriever calls made by the system to answer a single question, following \citet{DBLP:conf/naacl/JeongBCHP24}.
    \item \textbf{LM Calls (LMC)}: The average number of calls to the Language Model per question. Some systems may invoke the LM multiple times to calculate uncertainty, rephrase questions or generate additional rationales. 
\end{itemize}

\subsubsection{Self-Knowledge Metrics}

Self-knowledge is defined as a model's ability to recognize its own knowledge~\cite{yin2023large}. Measuring self-knowledge provides insight into the effectiveness of a method's adaptive retrieval component, as downstream performance is often influenced by external factors such as retriever selection, language model generation parameters, etc.

\begin{table*}[ht!]
    \centering
    \small
    \resizebox{\textwidth}{!}{

    \begin{tabular}{lccc|ccc|ccc|ccc|ccc|ccc}
    \toprule
    \multirow{2}{*}{{Method}} & \multicolumn{3}{c|}{NQ} & \multicolumn{3}{c|}{SQUAD} & \multicolumn{3}{c|}{TQA} & \multicolumn{3}{c|}{2Wiki} & \multicolumn{3}{c|}{HotPot} & \multicolumn{3}{c}{Musique} \\
    \cmidrule{2-19}
      & InAcc $\uparrow$ & LMC $\downarrow$ & RC $\downarrow$ & InAcc $\uparrow$ & LMC $\downarrow$ & RC $\downarrow$ & InAcc $\uparrow$ & LMC $\downarrow$ & RC $\downarrow$ & InAcc $\uparrow$ & LMC $\downarrow$ & RC $\downarrow$ & InAcc $\uparrow$ & LMC $\downarrow$ & RC $\downarrow$ & InAcc $\uparrow$ & LMC $\downarrow$ & RC $\downarrow$ \\
    \midrule
    Never RAG & 0.446 & 1.0 & 0.00 & 0.176 & 1.0 & 0.00 & 0.636 & 1.0 & 0.00 & 0.318 & 1.0 & 0.00 & 0.286 & 1.0 & 0.00 & 0.106 & 1.0 & 0.00 \\

    Always RAG & 0.496 & 1.0 & 1.00 & 0.312 & 1.0 & 1.00 & 0.610 & 1.0 & 1.00 & 0.374 & 1.0 & 1.00 & 0.410 & 1.0 & 1.00 & 0.100 & 1.0 & 1.00 \\
    \midrule
    IRCoT & 0.478 & 2.7 & 2.70 & 0.268 & 2.7 & 2.68 & 0.608 & 2.7 & 2.74 & 0.454 & 4.4 & 4.38 & \textbf{0.438} & 3.5 & 3.45 & 0.138 & 4.1 & 4.08 \\

    AdaptiveRAG & 0.496 & 2.0 & 0.98 & 0.286 & 2.0 & 0.97 & 0.628 & 1.5 & 0.54 & 0.454 & 5.2 & 2.64 & 0.414 & 4.6 & 2.34 & \textbf{0.140} & 3.6 & 3.63 \\
    DRAGIN & 0.480 & 4.5 & 2.24 & 0.298 & 4.3 & 2.14 & \textbf{0.666} & 4.1 & 2.06 & \textbf{0.456} & 5.8 & 2.92 & 0.430 & 5.1 & 2.56 & 0.134 & 6.3 & 3.15 \\

    FLARE & 0.450 & 3.1 & 2.07 & 0.238 & 3.1 & 2.08 & 0.648 & 2.1 & 1.39 & 0.424 & 3.9 & 2.85 & 0.372 & 5.1 & 4.07 & 0.090 & 4.1 & 3.10 \\

    Rowen\textsubscript{CL} & 0.494 & 29.5 & 7.24 & 0.196 & 29.2 & 7.19 & 0.656 & 28.7 & 7.06 & 0.444 & 32.9 & 7.87 & 0.354 & 31.9 & 7.67 & 0.104 & 42.1 & 9.52 \\
     Rowen\textsubscript{CM} & 0.494 & 29.5 & 7.27 & 0.196 & 29.2 & 7.20 & 0.656 & 28.7 & 7.12 & 0.444 & 32.9 & 7.87 & 0.356 & 31.9 & 7.70 & 0.104 & 42.1 & 9.52 \\
    Rowen\textsubscript{Hybrid} & 0.494 & 55.0 & 7.27 & 0.196 & 54.3 & 7.15 & 0.656 & 53.4 & 6.93 & 0.444 & 61.8 & 7.85 & 0.354 & 59.8 & 7.63 & 0.102 & 80.2 & 9.48 \\
    
    Seakr & 0.406 & 14.6 & 1.00 & 0.268 & 14.6 & 1.00 & 0.656 & 14.6 & 1.00 & 0.398 & 12.3 & 2.44 & 0.424 & 9.9 & 1.76 & 0.118 & 12.3 & 2.40 \\

    \midrule
    EigValLaplacian & \textbf{0.512} & 1.8 & 0.81 & 0.314 & 2.0 & 1.00 & 0.640 & 1.3 & 0.26 & \textbf{0.384} & 2.0 & 0.98 & 0.410 & 1.9 & 0.91 & \textbf{0.102} & 2.0 & \textbf{0.99} \\

    Lex-Similarity & \textbf{0.512} & 1.6 & \textbf{0.58} & \textbf{0.318} & 2.0 & \textbf{0.96} & 0.646 & 1.2 & 0.22 & 0.376 & 2.0 & 0.97 & 0.410 & 2.0 & 0.95 & 0.100 & 2.0 & 1.00 \\

    Max Entropy & 0.506 & 1.7 & 0.73 & 0.312 & 2.0 & 1.00 & 0.650 & 1.2 & \textbf{0.22} & 0.376 & 2.0 & 0.95 & \textbf{0.414} & 2.0 & 0.99 & 0.100 & 2.0 & 1.00 \\

    Mean Entropy & 0.498 & 1.9 & 0.88 & 0.314 & 2.0 & \textbf{0.95} & 0.650 & 1.3 & 0.30 & 0.378 & 1.9 & \textbf{0.93} & 0.410 & 2.0 & 0.99 & 0.100 & 2.0 & 1.00 \\

    SAR & 0.500 & 1.8 & 0.79 & 0.312 & 2.0 & 1.00 & 0.642 & 1.3 & 0.29 & 0.380 & 2.0 & 0.97 & 0.412 & 1.9 & \textbf{0.90} & 0.100 & 2.0 & 1.00 \\

    \midrule
    Best UE & \textbf{0.512} & 1.8 & 0.81 & \textbf{0.318} & 2.0 & 0.96 & \textbf{0.662} & 1.3 & 0.28 & \textbf{0.384} & 2.0 & 0.98 & \textbf{0.414} & 2.0 & 0.99 & \textbf{0.104} & 2.0 & \textbf{0.99} \\

    \midrule
    Ideal & 0.608 & 1.6 & 0.55 & 0.360 & 1.8 & 0.82 & 0.736 & 1.4 & 0.36 & 0.500 & 1.7 & 0.68 & 0.460 & 1.7 & 0.71 & 0.164 & 1.9 & 0.89 \\

    \bottomrule
    \end{tabular}
    }

    \caption{QA Performance of adaptive retrieval and uncertainty methods. `Best UE' refers to the top-performing uncertainty estimation method for each dataset. `Ideal' represents the performance of a system with an oracle providing ideal predictions for the need to retrieve. `InAcc' denotes In-Accuracy, measuring the QA system's performance. `LMC' indicates the mean number of LM calls per question, and `RC' represents the mean number of retrieval calls per question.}
    \label{tab:main_results}

\end{table*}

The task of identifying self-knowledge is formulated as a binary classification problem, where the ground truth label \( y \) is derived from the In-Accuracy of the model's response without external knowledge. Each method \( f \) can be represented as a function mapping input text \( x \) to a real-valued self-knowledge score \( f(x) \in \mathbb{R} \), where higher values indicate lower self-knowledge. The classification task is then performed by a classifier \( C \), producing the final prediction \( \hat{y} = C(f(x)) \in \{0, 1\} \).

For evaluation, we adopt metrics established in prior uncertainty estimation research~\cite{fadeeva2024fact, tao2024trust} and reflexive self-knowledge analysis~\cite{ni2024llms}.

\begin{itemize}[itemsep=0.3pt,topsep=0.5pt]
    \item \textbf{ROC-AUC (AUC)} evaluates the robustness of the method's self-knowledge identification performance: $\emph{AUC}(f(\mathbf{x}), \mathbf{y})$.
    \item \textbf{Spearman Correlation (Corr)} measures the alignment between the self-knowledge scores and the ground truth: $\emph{Corr}(f(\mathbf{x}), \mathbf{y})$.
    
    \item \textbf{Accuracy} quantifies the correctness of self-knowledge classifier: $\frac{1}{n}\sum\limits_{i=1}^n \mathbf{1}\{\hat{y_i}, y_i\}$.
    
    \item \textbf{Overconfidence} is a fraction of incorrect answers where the method was confident about self-knowledge reflecting how often the method incorrectly assumes that the model possesses the required knowledge when it does not: $\frac{1}{\sum\limits_{i} (1 - \hat{y_i})}\sum\limits_{i} (1-\mathbbm{1}\{\hat{y_i}, y_i\}) \cdot (1 - \hat{y_i})$.
    
    \item \textbf{Underconfidence} is a fraction of correct answers where the method was unconfident about self-knowledge reflecting how often the method fails to recognize that the model already has the required knowledge: $\frac{1}{\sum\limits_{i} \hat{y_i}}\sum\limits_{i} (1 - \mathbbm{1}\{\hat{y_i}, y_i\}) \cdot \hat{y_i}$.
\end{itemize}

\section{Results}

In the following sections, we describe the results of baseline and uncertainty methods for downstream performance, efficiency and self-knowledge. Along with the end-to-end and UE methods, we also apply two additional methods for better comparison. ``Best UE'' refers to the top-performing
uncertainty estimation method for each dataset. ``Ideal'' represents the performance of a system with an oracle
providing ideal predictions for the need to retrieve.

\subsection{Downstream and Efficiency Performance}

The results in Table~\ref{tab:main_results} show that uncertainty estimation methods outperform baseline methods on single-hop datasets and perform comparably on multi-hop datasets, while being significantly more compute-efficient, often several times cheaper.

While baseline methods may achieve slightly better performance on some datasets, they require multiple calls to both the language model and retriever, leading to higher computational costs. In contrast, uncertainty estimation methods consistently require fewer than one retriever call and two or less LM calls per question, significantly reducing inference costs.

Uncertainty estimation for adaptive retrieval consistently outperforms constant retrieval in terms of performance and efficiency. However, analysis of the Ideal uncertainty estimator reveals that current methods still fall short of perfect performance, both in terms of efficiency and In-Accuracy, highlighting the ongoing challenge of accurately identifying self-knowledge within the model.

\begin{table*}[t]
    \centering
    \small
    \resizebox{\textwidth}{!}{

\begin{tabular}{lccc|ccc|ccc|ccc|ccc|ccc}
\toprule
\multirow{2}{*}{{Method}} & \multicolumn{3}{c|}{NQ} & \multicolumn{3}{c|}{SQUAD} & \multicolumn{3}{c|}{TQA} & \multicolumn{3}{c|}{2Wiki} & \multicolumn{3}{c|}{HotPot} & \multicolumn{3}{c}{Musique} \\
\cmidrule{2-19}
 & Acc & Corr & AUC & Acc & Corr & AUC & Acc & Corr & AUC & Acc & Corr & AUC & Acc & Corr & AUC & Acc & Corr & AUC \\
\midrule
AdaptiveRAG & 0.57 & 0.06 & 0.54 & 0.73 & 0.10 & 0.58 & 0.51 & -0.02 & 0.49 & \textbf{0.72} & \textbf{0.34} & \textbf{0.71} & 0.71 & 0.19 & 0.62 & 0.88 & 0.15 & 0.64 \\
DRAGIN & 0.55 & 0.12 & 0.57 & 0.82 & 0.11 & 0.58 & 0.36 & 0.03 & 0.52 & 0.68 & -0.07 & 0.46 & 0.71 & 0.01 & 0.51 & \underline{0.89} & -0.01 & 0.49 \\
FLARE & 0.59 & 0.16 & 0.59 & 0.54 & 0.11 & 0.58 & 0.58 & 0.12 & 0.57 & 0.51 & 0.20 & 0.62 & 0.42 & 0.06 & 0.54 & 0.59 & 0.01 & 0.51 \\
Rowen\textsubscript{CL} & 0.45 & -0.14 & 0.44 & 0.18 & -0.06 & 0.47 & 0.64 & -0.07 & 0.47 & 0.32 & -0.10 & 0.46 & 0.29 & -0.13 & 0.44 & 0.11 & 0.00 & 0.50 \\
Rowen\textsubscript{CM} & 0.45 & -0.03 & 0.49 & 0.18 & -0.06 & 0.47 & 0.64 & -0.13 & 0.44 & 0.32 & 0.02 & 0.51 & 0.29 & -0.14 & 0.44 & 0.11 & -0.02 & 0.49 \\
Rowen\textsubscript{Hybrid} & 0.45 & -0.12 & 0.44 & 0.17 & -0.07 & 0.46 & 0.63 & -0.13 & 0.43 & 0.32 & -0.04 & 0.48 & 0.29 & -0.17 & 0.41 & 0.11 & -0.01 & 0.49 \\
Seakr & 0.55 & \textbf{0.24} & \textbf{0.64} & 0.82 & \textbf{0.36} & \textbf{0.77} & 0.36 & \textbf{0.47} & \textbf{0.78} & 0.68 & -0.22 & 0.37 & 0.71 & 0.08 & 0.55 & \underline{0.89} & 0.06 & 0.56 \\
\midrule
EigValLaplacian & 0.60 & 0.17 & 0.60 & 0.83 & 0.10 & 0.57 & 0.70 & 0.34 & 0.71 & \underline{0.69} & 0.19 & 0.62 & \textbf{0.73} & 0.27 & 0.67 & \underline{0.89} & 0.12 & 0.62 \\

Lex-Similarity & \underline{0.61} & 0.22 & \underline{0.63} & \textbf{0.84} & 0.22 & 0.67 & \textbf{0.73} & \underline{0.39} & \underline{0.74} & 0.68 & 0.21 & 0.63 & \textbf{0.73} & 0.30 & 0.69 & \textbf{0.90} & 0.08 & 0.59 \\
Max Entropy & \textbf{0.63} & 0.20 & 0.62 & 0.82 & 0.25 & \underline{0.69} & 0.72 & 0.35 & 0.71 & 0.69 & 0.19 & 0.62 & \textbf{0.73} & 0.29 & 0.69 & \underline{0.89} & \textbf{0.18} & \textbf{0.67} \\
Mean Entropy & \underline{0.61} & 0.20 & 0.62 & \textbf{0.84} & \underline{0.32} & \underline{0.74} & 0.72 & 0.36 & 0.72 & 0.68 & \underline{0.28} & \underline{0.68} & \underline{0.72} & \textbf{0.31} & \textbf{0.70} & \textbf{0.90} & 0.15 & 0.64 \\
SAR & \underline{0.61} & \underline{0.23} & \underline{0.63} & \underline{0.83} & 0.28 & 0.71 & 0.72 & 0.38 & 0.73 & \underline{0.69} & 0.23 & 0.65 & \textbf{0.73} & 0.30 & 0.69 & \underline{0.89} & \underline{0.17} & \underline{0.66} \\
\bottomrule
\end{tabular}
}
    \caption{\textbf{Self-knowledge} metrics for adaptive retrieval and uncertainty methods. `Acc' and `AUC' refer to accuracy and ROC-AUC, respectively, for identifying self-knowledge. `Corr' denotes the Spearman correlation with the self-knowledge label. Bold values indicate the highest score,  underlined values represent the second-highest score.}
    \label{tab:main_proxy}
    \vspace{-0.2cm}
\end{table*}

\begin{tcolorbox}[colback=gray!3, colframe=gray!50]
\textbf{Takeaway 1:} Uncertainty methods outperform baselines on single-hop tasks, match them on multi-hop tasks, and are far more efficient.  The ``Ideal'' estimator highlights room for improvement in the self-knowledge identification.

\end{tcolorbox}

\subsection{Self-Knowledge Performance}

The results in Table~\ref{tab:main_proxy} demonstrate that, despite strong downstream performance, most adaptive retrieval methods may lack the ability to accurately identify self-knowledge, exhibiting near-zero correlation and random predictions. For instance, while DRAGIN typically dominates on downstream tasks, it performs poorly on self-knowledge metrics.

In contrast, SeaKR exhibits strong self-knowledge identification on single-hop datasets, underscoring the value of inspecting the internal states of language models. However, SeaKR's performance declines on multi-hop datasets, where internal states may provide limited information about the model's knowledge of more complex questions. For multi-hop tasks, AdaptiveRAG demonstrates superior results, highlighting the effectiveness of reflexive trainable methods, which apparently handle complex reasoning better.

{These results suggest that internal-state uncertainty excels for simple questions, while reflexive uncertainty methods are better suited for complex reasoning tasks.}

\begin{figure}[t]
    \centering
    \includegraphics[width=\linewidth]{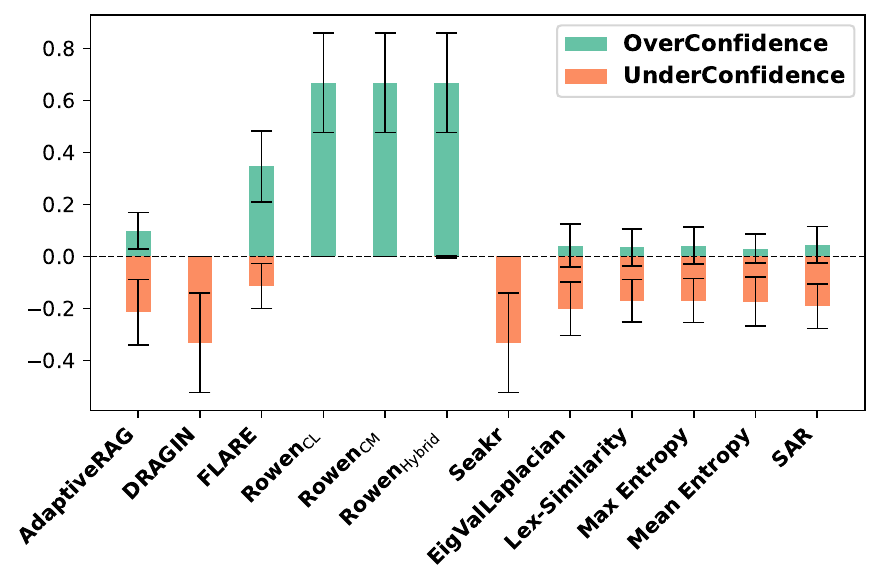}
    \caption{Average overconfidence and underconfidence for each method. Deviation from the zero value is undesirable and indicates erroneous behavior. High \textbf{OverConfidence} values reflect cases where the method incorrectly assumes the model has the required knowledge when it does not. High \textbf{UnderConfidence} values indicate instances where the method fails to recognize that the model already possesses the required knowledge.
    }
    \label{fig:confidence}
    \vspace{-0.2cm}
\end{figure}

According to the results in Figure~\ref{fig:confidence}, nearly all baseline models, except for AdaptiveRAG, exhibit a tendency to either consistently overestimate self-knowledge or, conversely, to be underconfident. In contrast, uncertainty methods strike the best balance between overconfidence and underconfidence, demonstrating more adequate and reliable values.

Overall, uncertainty estimation methods consistently exhibit the strongest ability to identify self-knowledge, ranking first or second across all methods. These findings emphasize the need for a more thorough evaluation of adaptive retrieval methods, beyond relying solely on downstream performance, showing no significant correlation, further shown in Table~\ref{tab:metric_corr}.

\begin{tcolorbox}[colback=gray!3, colframe=gray!50]
\textbf{Takeaway 2:} Internal-based SeaKR excels at simple tasks, while reflexive AdaptiveRAG performs better on complex ones. Uncertainty methods provide the most reliable self-knowledge estimates, emphasizing evaluation beyond QA performance.

\end{tcolorbox}

\begin{figure*}[t]
    \centering
    \includegraphics[width=0.95\linewidth]{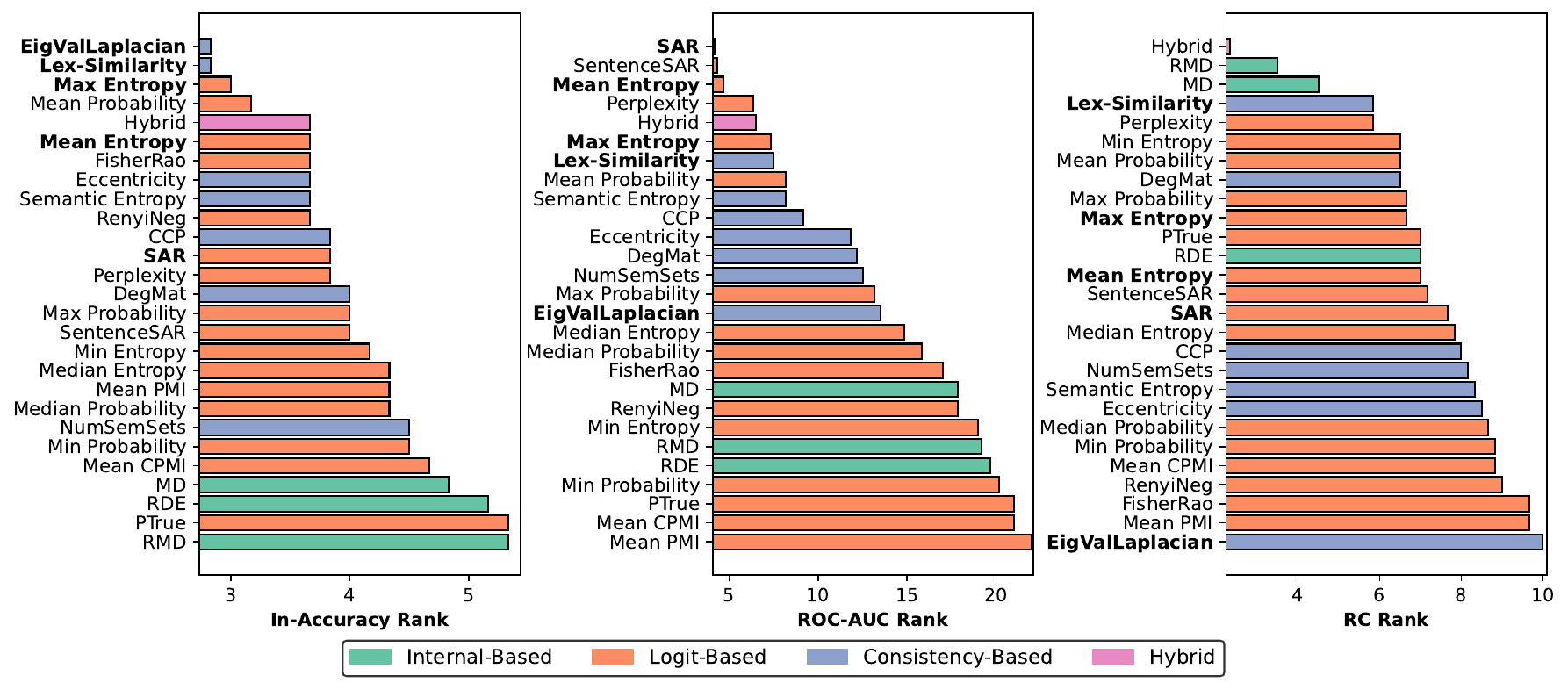}
    \caption{Uncertainty methods average ranks for In-Accuracy, ROC-AUC and Retrieval Calls. Smaller rank indicate average better performance. The In-Accuracy ranks demonstrate key downstream metrics, while the ROC-AUC ranks show self-knowledge abilities across different methods, affecting average downstream performance. The Retriever Calls~(RC) ranks represent the efficiency of the method. This evaluation led to choose \textbf{EigValLaplacian}, \textbf{Lex-Similarity}, \textbf{Max Entropy}, \textbf{Mean Entropy}, and \textbf{SAR} for more detailed analysis.}
    \label{fig:uc_ranks}
    \vspace{-0.2cm}
\end{figure*}

\begin{figure*}[t]
    \centering
    \includegraphics[width=0.95\linewidth]{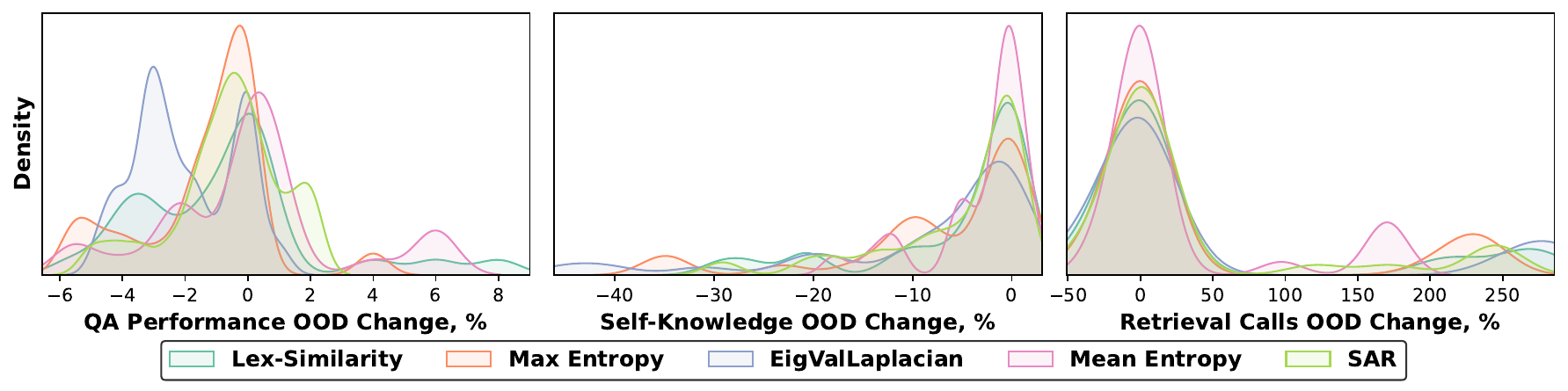}
    \caption{The transferability of methods between datasets was evaluated using average changes in metrics for Out-Of-Distribution~(OOD) data. QA~Performance in OOD was measured by InAccuracy, showing comparable results across methods. Self-Knowledge, evaluated by Accuracy, degraded significantly. Efficiency was assessed by RC, indicating that methods tend to call the retriever more frequently after transfer.}
    \label{fig:uc_ood}
    \vspace{-0.2cm}
\end{figure*}

\subsection{Uncertainty Estimation}

We analyze 27 uncertainty estimation methods across QA performance, efficiency, and self-knowledge, categorizing them by underlying approach. Methods are ranked based on their average performance across datasets, with smaller ranks indicating better results.

As shown in Figure~\ref{fig:uc_ranks}, EigValLaplacian and Lex-Similarity rank highest for In-Accuracy, while SAR variants and Mean Entropy dominate for ROC-AUC, highlighting an inconsistency between self-knowledge and downstream performance. This discrepancy is further evidenced by a moderate Spearman correlation of $0.65$ between In-Accuracy and ROC-AUC ranks, likely due to differing sensitivities to Type I and II errors. EigValLaplacian also ranks highest for Retrieval Calls, indicating overconfidence.

For our main analysis, we select uncertainty methods with the best QA performance: EigValLaplacian, Lex-Similarity, and Max Entropy and top self-knowledge methods: SAR and Mean Entropy for self-knowledge assessment. Internal-state methods generally rank lower for In-Accuracy and ROC-AUC but perform better in efficiency, suggesting overconfidence. Consistency-based methods excel in QA performance but drop in self-knowledge, lagging behind logit-based methods, indicating better stability to distribution shifts.

The Hybrid method balances all metrics, ranking in the top-5 for In-Accuracy and ROC-AUC and first for efficiency. However, it requires calculating all uncertainty estimates, introducing computational overhead, which may still be justified in retrieval-limited scenarios.

Finally, we analyze feature importance for the Hybrid method in details in Figure~\ref{fig:fi_heatmap} in Appendix~\ref{appx:fi_hybrid_ue}.

\begin{tcolorbox}[colback=gray!3, colframe=gray!50]

\textbf{Takeaway 3:} Consistency-based methods excel in downstream performance but lag in self-knowledge, while logit-based methods dominate self-knowledge metrics. The Hybrid method balances all metrics but incurs higher computational costs.
\end{tcolorbox}

\section{Out-of-Domain Transfer}

To analyze the robustness of UE methods on out-of-domain (OOD) datasets, we evaluate their performance across all possible dataset pairs by training on each dataset and testing on every other. The relative change in performance, expressed as a percentage compared to in-domain performance (see Appendix \ref{sec:appendix_ood_heatmap}), is used to assess OOD robustness. For statistical tests details, refer to Appendix~\ref{sec:ood_stat}.

In Figure~\ref{fig:uc_ood}, we present the distributions of performance change across all train-test dataset pairs. For In-Accuracy, most methods perform comparably, with EigValLaplacian being the only method that significantly lags behind and differs from nearly all others. While most methods are centered around 0, indicating stability, there is a noticeable tail representing a loss in quality. Nevertheless, the loss typically remains under 4\%, suggesting strong downstream OOD transfer performance, with occasional cases of positive improvement.

However, the QA performance can be influenced by multiple factors. Self-knowledge transfer, measured by Accuracy, reveals a complex picture. While the changes are centered around 0—an encouraging sign of stability—the tail indicating quality loss is notably heavier, with more extreme variations and no cases of positive transfer. EigValLaplacian stands out with the weakest transfer performance, whereas other methods show comparable results without statistically significant differences.

Efficiency transfer analysis shows a similar centering around 0 but reveals the largest percentage changes. Methods tend to call the retriever more frequently when transferred, indicating underconfidence. No significant differences are observed between methods.

\begin{tcolorbox}[colback=gray!3, colframe=gray!50]
\textbf{Takeaway 4:} UE methods show strong OOD robustness for QA performance but lower for self-knowledge and efficiency, with no significant differences between most methods.
\end{tcolorbox}

\section{Uncertainty Estimation Complexity}

\begin{figure}[t!]
    \centering
    \includegraphics[width=\linewidth]{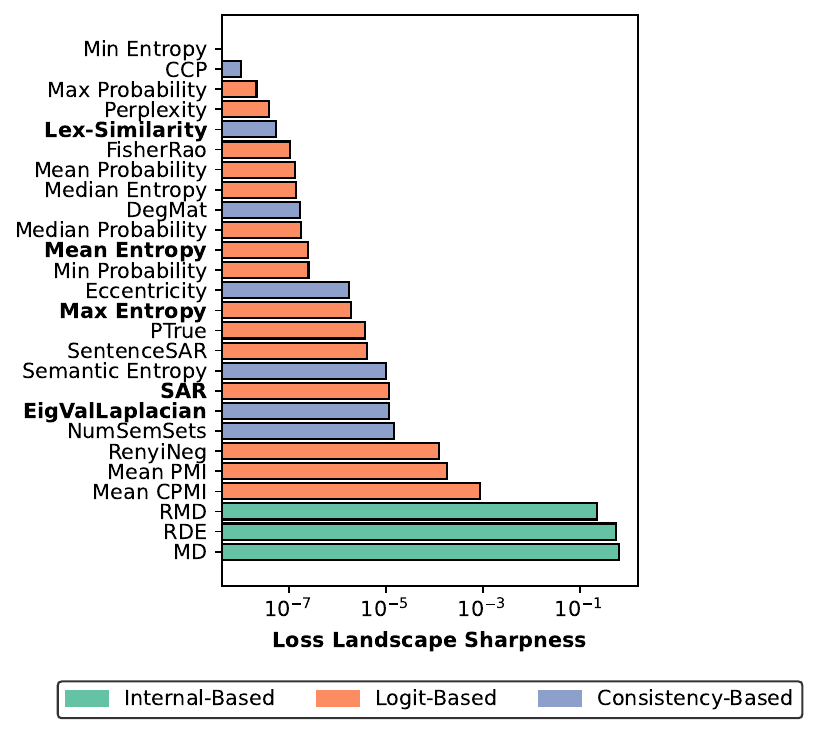}
    \caption{Average loss landscape sharpness in logarithmic scale. Higher values correspond to more complex functions.}
    \label{fig:sharpness}
    \vspace{-0.2cm}
\end{figure}



We analyze the complexity of uncertainty estimation methods $f$ with logistic regression $C(f)$, enabling rigorous evaluation of \( f \)'s complexity.

To achieve so, we employ Rademacher Complexity for functional complexity~\cite{yin2019rademacher} and sharpness of loss landscape with Hessian eigenvalues~\cite{sagun2016eigenvalues,pmlr-v9-glorot10a}.

\paragraph{Rademacher Complexity} quantifies the capacity of a hypothesis class to fit random noise with higher values indicating greater complexity:
\[
\mathcal{R}_n(\mathcal{H}_f) = \mathbb{E}_{\sigma} \left[ \sup_{h \in \mathcal{H}_f} \frac{1}{n} \sum_{i=1}^n \sigma_i h(x_i) \right],
\]
where \( \mathcal{H}_f \) is the hypothesis class induced by uncertainty method \( f \), \( \sigma_i \sim \mathbb{U}\{-1, 1\} \) are Rademacher random variables, and \( h(x_i) \) is the model's prediction.

\paragraph{Loss Landscape Sharpness} quantifies complexity from different perspective evaluating the curvature of loss landscape, with higher values indicating more complex and harder generalizable functions~\cite{pmlr-v187-kaur23a}. 

Let \( \mathcal{L}(w) \in C^2 \) be a twice continuously differentiable loss function with respect to \( w \in \mathbb{R}^d \), and let \( H(w) = \nabla^2_w \mathcal{L}(w) \) denote its Hessian. The sharpness at the optimized parameters \( w^* \) is defined as:
\[
\lambda_{\max} = \sup_{\|v\|_2 = 1} v^\top H(w^*) v,
\]
where \( \lambda_{\max} \) is the largest eigenvalue of \( H(w^*) \), capturing the steepest curvature of the loss surface. 

\begin{figure}[t!]
    \centering
    \includegraphics[width=\linewidth]{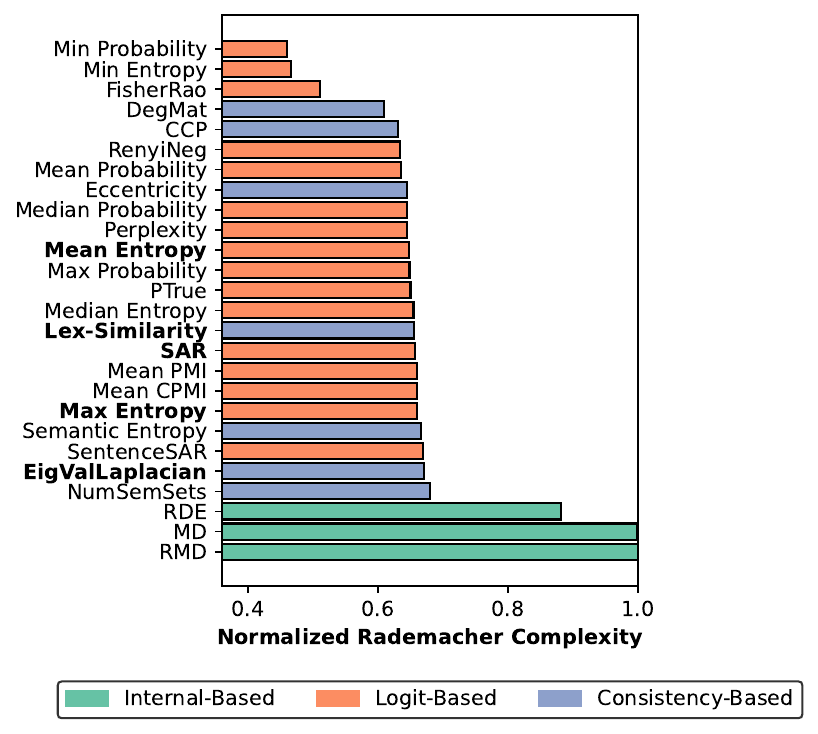}
    \caption{Normalized Rademacher Complexity for uncertainty methods. Higher values indicate richer complexity of feature.}
    \label{fig:rademacher}
    \vspace{-0.2cm}
\end{figure}

Figures~\ref{fig:sharpness} and~\ref{fig:rademacher} show that internal-based features induce the most complex functions with sharper loss landscapes. In contrast, moderately complex features like EigValLaplacian achieve better QA performance, while overall consistency-based methods are more complex than logit-based ones.

\begin{tcolorbox}[colback=gray!3, colframe=gray!50]
\textbf{Takeaway 5:} Internal-based methods are the most complex and harder to generalize, while consistency-based methods are more complex than logit-based ones.
\end{tcolorbox}

\section{Conclusion}

We present a comprehensive computational study of adaptive retrieval systems, evaluating 27 established uncertainty estimation methods alongside 8 recently published methods tailored for this task. Our analysis considers downstream QA performance, efficiency, and self-knowledge, covering a total of 10 evaluation metrics.
Our findings show that established uncertainty methods achieve performance comparable to recently proposed adaptive retrieval approaches, while being more efficient and exhibiting stronger self-knowledge capabilities. 

Moreover, we conducted an in-depth comparison of the 27 uncertainty estimation methods, revealing notable discrepancies between downstream performance and self-knowledge metrics. Our analysis of OOD transfer shows minimal deviations in downstream performance but a significant decline in self-knowledge, with no substantial differences observed between methods. We also identify the higher functional complexity of internal-based methods.




\section*{Limitations}

\begin{itemize} 
    \item We conduct our study using the LLaMA3.1-8b-instruct model, which is among the best open-source models within its parameter range. However, extending the analysis to additional models would help validate the consistency of our findings across different architectures. 
    \item Our evaluation is performed on 6 QA datasets, which are standard for this task. Expanding the evaluation to include more QA datasets, particularly domain-specific ones, could uncover additional insights and highlight the generalizability of the methods. \end{itemize}

\section*{Ethical Considerations}

Text information retrieval systems may yield biased text documents, biasing the resulting generation  of even an aligned ethically LLM in an undesired direction. Therefore, engineers deploying RAG and Adaptive RAG pipelines in real world applications facing users shall consider this potential issue.

\bibliography{preprint}

\appendix



\clearpage
\onecolumn

\section{Classifier for UE} \label{sec:classifier_sens}

We further analyze how the choice of classifier (Logistic Regression, Threshold, KNN, MLP or Decision Tree) algorithm impacts QA performance. Specifically, we compute the average performance drop for each uncertainty method when switching from the maximum classifier performance to the average. This sensitivity further indicates the complexity of the method, as more complex methods require thorough choice of classifier and hyperparameters.

\begin{figure}[ht!]
    \centering
    \includegraphics[width=0.6\linewidth]{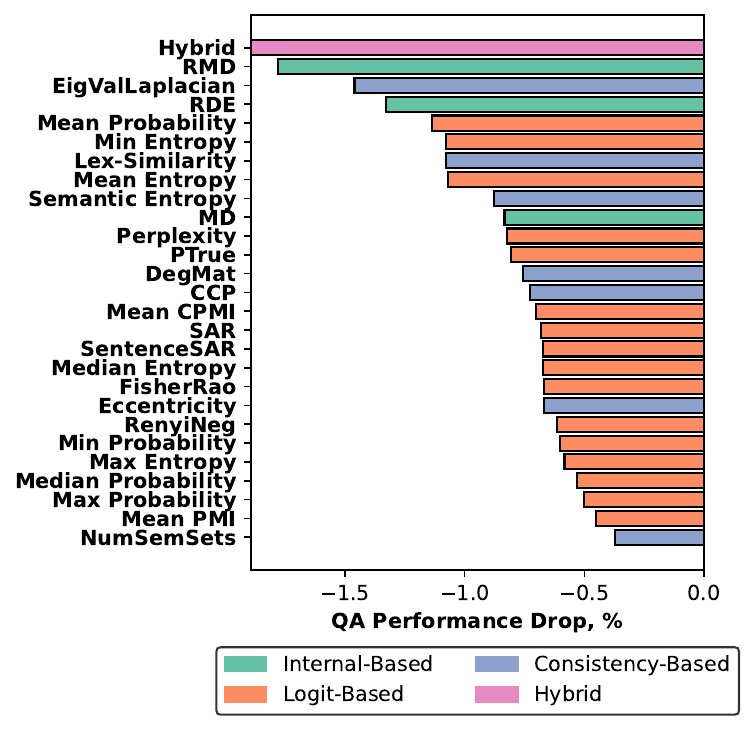}
    \caption{Average QA performance drop for uncertainty methods for when switching maximum over classifiers to average.}
    \label{fig:classifier_sens}
\end{figure}

The results in Figure~\ref{fig:classifier_sens} demonstrate that, except for NumSemSets (the number of semantic clusters of sampled responses), consistency-based methods exhibit higher sensitivity compared to logit-based methods. Internal state-based methods and hybrid approaches show the greatest sensitivity, which aligns with their complex nature. This increased sensitivity likely arises because their features are less explicit, capturing subtle internal state changes that are inherently harder to fit and explain.

We further illustrate the ranking differences in Table~\ref{tab:classifier_sens}, showing that methods with greater stability tend to achieve higher ranks when averaged across classifiers.

\begin{table*}[t]
    \centering
    \small
\begin{tabular}{lccc}
\toprule
Method & Mean & Max & Difference \\
\midrule
Hybrid & 25.67 & 11.33 & -14.33 \\
RMD & 23.50 & 15.00 & -8.50 \\
Perplexity & 15.00 & 8.83 & -6.17 \\
MD & 19.50 & 13.50 & -6.00 \\
CCP & 15.50 & 9.67 & -5.83 \\
EigValLaplacian & 11.67 & 6.00 & -5.67 \\
Median Entropy & 15.67 & 10.67 & -5.00 \\
Mean Entropy & 12.83 & 8.33 & -4.50 \\
RDE & 16.83 & 12.67 & -4.17 \\
DegMat & 13.67 & 10.00 & -3.67 \\
Max Entropy & 10.00 & 6.50 & -3.50 \\
Mean CPMI & 14.00 & 11.00 & -3.00 \\
Lex-Similarity & 9.33 & 6.50 & -2.83 \\
SentenceSAR & 12.00 & 9.17 & -2.83 \\
Min Entropy & 11.83 & 9.33 & -2.50 \\
SAR & 10.50 & 8.83 & -1.67 \\
PTrue & 16.00 & 14.50 & -1.50 \\
Max Probability & 11.33 & 10.50 & -0.83 \\
Min Probability & 11.00 & 10.17 & -0.83 \\
FisherRao & 9.17 & 8.50 & -0.67 \\
Semantic Entropy & 9.17 & 8.50 & -0.67 \\
RenyiNeg & 9.00 & 9.33 & 0.33 \\
Mean Probability & 5.83 & 6.33 & 0.50 \\
Mean PMI & 10.33 & 11.17 & 0.83 \\
Median Probability & 8.83 & 9.83 & 1.00 \\
Eccentricity & 7.83 & 8.83 & 1.00 \\
NumSemSets & 10.33 & 12.00 & 1.67 \\
\bottomrule
\end{tabular}
   \caption{Rank of UC methods by In-Accuracy, aggregated using the mean or maximum across different classifiers. A lower difference indicates reduced stability to classifier choice, whereas a higher difference reflects greater robustness to classifier choice.
}
    \label{tab:classifier_sens}
\end{table*}


\section{Statistical Tests for OOD Testing} \label{sec:ood_stat}

To evaluate the OOD performance differences between uncertainty estimation methods under dataset-dependent conditions, we use the Friedman test, suitable for data with small sample sizes and no assumptions about normality, while also being appropriate for repeated measurements.

After the Friedman test, we apply the Nemenyi post-hoc test to identify statistically significant pairwise differences between methods, similarly due to rank-based nature and accounting for multiple comparisons to ensure robust analysis. We also report significance with asterisk atop of the number.

\begin{table*}[th!]
    \centering
    \small
    \resizebox{\textwidth}{!}{
    
\begin{tabular}{lcccccccccccc}
\toprule
\multirow{2}{*}{{Method}} & \multicolumn{2}{c}{NQ} & \multicolumn{2}{c}{SQUAD} & \multicolumn{2}{c}{TQA} & \multicolumn{2}{c}{2Wiki} & \multicolumn{2}{c}{HotPot} & \multicolumn{2}{c}{Musique} \\
 & Over & Under & Over & Under & Over & Under & Over & Under & Over & Under & Over & Under \\
\midrule
AdaptiveRAG & 0.01 & 0.43 & 0.14 & 0.13 & 0.19 & 0.30 & 0.12 & 0.15 & 0.11 & 0.18 & 0.02 & 0.10 \\
DRAGIN & 0.00 & 0.45 & 0.00 & 0.18 & 0.00 & 0.64 & 0.00 & 0.32 & 0.00 & 0.29 & 0.00 & 0.11 \\

FLARE & 0.20 & 0.21 & 0.40 & 0.06 & 0.18 & 0.24 & 0.43 & 0.06 & 0.53 & 0.05 & 0.34 & 0.06 \\

Rowen\textsubscript{CL} & 0.55 & 0.00 & 0.82 & 0.00 & 0.36 & 0.00 & 0.68 & 0.00 & 0.71 & 0.00 & 0.89 & 0.00 \\
Rowen\textsubscript{CM} & 0.55 & 0.00 & 0.82 & 0.00 & 0.36 & 0.00 & 0.68 & 0.00 & 0.71 & 0.00 & 0.89 & 0.00 \\
Rowen\textsubscript{Hybrid} & 0.55 & 0.00 & 0.82 & 0.00 & 0.36 & 0.01 & 0.68 & 0.00 & 0.71 & 0.00 & 0.89 & 0.00 \\

Seakr & 0.00 & 0.45 & 0.00 & 0.18 & 0.00 & 0.64 & 0.00 & 0.32 & 0.00 & 0.29 & 0.00 & 0.11 \\
\midrule
Lex-Similarity & 0.01 & 0.21 & 0.00 & 0.15 & 0.18 & 0.06 & 0.00 & 0.28 & 0.02 & 0.22 & 0.00 & 0.10 \\
Max Entropy & 0.08 & 0.20 & 0.00 & 0.13 & 0.17 & 0.07 & 0.00 & 0.29 & 0.00 & 0.23 & 0.00 & 0.10 \\
EigValLaplacian & 0.03 & 0.33 & 0.00 & 0.17 & 0.21 & 0.08 & 0.00 & 0.30 & 0.01 & 0.23 & 0.00 & 0.10 \\
SAR & 0.06 & 0.29 & 0.00 & 0.16 & 0.18 & 0.09 & 0.00 & 0.28 & 0.03 & 0.22 & 0.00 & 0.11 \\
Mean Entropy & 0.04 & 0.29 & 0.00 & 0.14 & 0.14 & 0.07 & 0.00 & 0.29 & 0.00 & 0.15 & 0.00 & 0.10 \\
\bottomrule
\end{tabular}

}

    \caption{Over- and UnderConfidence for adaptive retrieval methods and uncertainty estimation. Values closest to zero indicate the best performance. UnderConfidence refers to cases where the method failed to detect self-knowledge despite its presence, while OverConfidence reflects cases where the method incorrectly detected self-knowledge when it was absent.}
    \label{tab:confidence}
\end{table*}


\begin{table*}[t]
    \centering
    \small
    \resizebox{\textwidth}{!}{

\begin{tabular}{ccccc|cccc|cccc}
\toprule
\multirow{2}{*}{{Method}}  & \multicolumn{4}{c|}{NQ} & \multicolumn{4}{c|}{SQUAD} & \multicolumn{4}{c}{TQA} \\
 & EM & F1 & InAcc & RC & EM & F1 & InAcc & RC & EM & F1 & InAcc & RC \\
\midrule
CCP & 0.398 & 0.512 & 0.496 & 0.94 & 0.252 & 0.389 & 0.312 & 1.00 & \textbf{0.600} & \textbf{0.692} & \textbf{0.662} & 0.28 \\
DegMat & 0.394 & 0.514 & 0.496 & 0.97 & 0.252 & 0.389 & 0.312 & 1.00 & 0.598 & 0.684 & 0.644 & 0.29 \\
Eccentricity & 0.404 & 0.520 & 0.500 & 0.84 & 0.252 & 0.390 & 0.312 & 1.00 & 0.594 & 0.677 & 0.638 & 0.21 \\
EigValLaplacian & 0.406 & 0.532 & \textbf{0.512} & 0.81 & 0.254 & 0.391 & 0.314 & 1.00 & 0.594 & 0.682 & 0.640 & 0.26 \\
FisherRao & 0.390 & 0.506 & 0.498 & 0.88 & 0.252 & 0.389 & 0.312 & 1.00 & 0.598 & 0.688 & 0.654 & 0.11 \\
Hybrid & 0.410 & 0.534 & 0.504 & 0.65 & 0.254 & 0.393 & 0.314 & 0.99 & 0.594 & 0.689 & 0.654 & 0.32 \\
Lex-Similarity & 0.420 & \textbf{0.535} & \textbf{0.512} & 0.58 & 0.256 & \textbf{0.394} & \textbf{0.318} & 0.96 & \textbf{0.600} & 0.689 & 0.646 & 0.22 \\
MD & 0.398 & 0.511 & 0.496 & 1.00 & 0.252 & 0.389 & 0.312 & 1.00 & 0.598 & 0.681 & 0.642 & 0.05 \\
Max Entropy & \textbf{0.422} & \textbf{0.535} & 0.506 & 0.73 & 0.252 & 0.389 & 0.312 & 1.00 & 0.598 & 0.689 & 0.650 & 0.22 \\
Max Probability & 0.418 & 0.532 & 0.502 & 0.82 & 0.252 & 0.389 & 0.312 & 1.00 & 0.592 & 0.683 & 0.646 & 0.21 \\
Mean CPMI & 0.390 & 0.506 & 0.496 & 1.00 & 0.252 & 0.389 & 0.312 & 1.00 & 0.592 & 0.675 & 0.640 & 0.02 \\
Mean Entropy & 0.402 & 0.514 & 0.498 & 0.88 & 0.254 & 0.392 & 0.314 & 0.95 & 0.598 & 0.687 & 0.650 & 0.30 \\
Mean PMI & 0.390 & 0.506 & 0.496 & 1.00 & 0.254 & 0.389 & 0.312 & 1.00 & 0.596 & 0.683 & 0.640 & 0.02 \\
Mean Probability & 0.404 & 0.512 & 0.498 & 0.77 & 0.258 & \textbf{0.394} & \textbf{0.318} & 0.98 & 0.592 & 0.681 & 0.642 & 0.06 \\
Median Entropy & 0.412 & 0.519 & 0.496 & 1.00 & 0.252 & 0.389 & 0.312 & 1.00 & 0.596 & 0.682 & 0.644 & 0.15 \\
Median Probability & 0.408 & 0.512 & 0.496 & 1.00 & 0.252 & 0.389 & 0.312 & 1.00 & 0.592 & 0.680 & 0.644 & 0.26 \\
Min Entropy & 0.398 & 0.515 & 0.504 & 0.93 & 0.252 & 0.389 & 0.312 & 1.00 & 0.592 & 0.675 & 0.636 & 0.00 \\
Min Probability & 0.398 & 0.515 & 0.502 & 0.91 & 0.252 & 0.389 & 0.312 & 1.00 & 0.592 & 0.675 & 0.636 & 0.00 \\
NumSemSets & 0.406 & 0.521 & 0.502 & 0.83 & 0.252 & 0.389 & 0.312 & 1.00 & 0.590 & 0.680 & 0.638 & 0.28 \\
PTrue & 0.388 & 0.506 & 0.496 & 1.00 & 0.252 & 0.389 & 0.312 & 1.00 & 0.592 & 0.676 & 0.636 & 0.00 \\
Perplexity & 0.404 & 0.515 & 0.498 & 0.77 & 0.256 & 0.392 & 0.316 & 0.98 & 0.594 & 0.683 & 0.646 & 0.16 \\
RDE & 0.388 & 0.506 & 0.496 & 1.00 & 0.252 & 0.389 & 0.312 & 1.00 & 0.588 & 0.670 & 0.634 & 0.08 \\
RMD & 0.394 & 0.508 & 0.496 & 1.00 & 0.252 & 0.389 & 0.312 & 1.00 & 0.592 & 0.675 & 0.636 & 0.00 \\
RenyiNeg & 0.402 & 0.517 & 0.498 & 0.96 & 0.252 & 0.389 & 0.312 & 1.00 & 0.594 & 0.688 & 0.654 & 0.24 \\
SAR & 0.410 & 0.526 & 0.500 & 0.79 & 0.254 & 0.389 & 0.312 & 1.00 & 0.590 & 0.681 & 0.642 & 0.29 \\
Semantic Entropy & 0.406 & 0.521 & 0.504 & 0.83 & \textbf{0.260} & 0.393 & 0.316 & 0.92 & 0.596 & 0.685 & 0.640 & 0.24 \\
SentenceSAR & 0.410 & 0.521 & 0.500 & 0.73 & 0.254 & 0.391 & 0.314 & 0.99 & 0.596 & 0.685 & 0.644 & 0.24 \\
\bottomrule
\end{tabular}
    }
    \caption{Detailed QA performance results for uncertainty methods on one-hop datasets. `InAcc' denotes In-Accuracy, and `EM' stands for Exact Match. Higher values indicate better performance. Bold values highlight the best results. Standard deviations for InAcc, EM, and F1 are $\approx 0.02 \pm 0.003$, calculated using bootstrapping.}
    \label{tab:ue_result_one_hop}
\end{table*}

\begin{table*}[t]
    \centering
    \small
    \resizebox{\textwidth}{!}{
\begin{tabular}{ccccc|cccc|cccc}
\toprule
\multirow{2}{*}{{Method}}  & \multicolumn{4}{c|}{2Wiki} & \multicolumn{4}{c|}{HotPot} & \multicolumn{4}{c}{Musique} \\
 & EM & F1 & InAcc & RC & EM & F1 & InAcc & RC & EM & F1 & InAcc & RC \\
\midrule




CCP & 0.310 & 0.398 & 0.376 & 0.98 & 0.386 & 0.497 & 0.410 & 1.00 & 0.088 & 0.167 & 0.100 & 1.00 \\
DegMat & \textbf{0.314} & \textbf{0.407} & 0.382 & 0.95 & 0.386 & 0.498 & 0.410 & 1.00 & 0.088 & 0.168 & 0.100 & 1.00 \\
Eccentricity & 0.312 & 0.406 & \textbf{0.384} & 0.93 & \textbf{0.390} & \textbf{0.502} & \textbf{0.414} & 0.93 & 0.088 & 0.167 & 0.100 & 1.00 \\
EigValLaplacian & 0.312 & 0.405 & \textbf{0.384} & 0.98 & 0.384 & 0.501 & 0.410 & 0.91 & 0.088 & 0.169 & 0.102 & 1.00 \\
FisherRao & 0.306 & 0.399 & 0.378 & 0.98 & 0.386 & 0.497 & 0.410 & 1.00 & 0.088 & 0.169 & 0.100 & 1.00 \\
Hybrid & 0.298 & 0.391 & 0.368 & 0.93 & 0.384 & 0.491 & 0.406 & 0.94 & 0.090 & 0.169 & 0.102 & 1.00 \\
Lex-Similarity & 0.306 & 0.400 & 0.376 & 0.97 & 0.386 & 0.498 & 0.410 & 0.95 & 0.088 & 0.168 & 0.100 & 1.00 \\
MD & 0.302 & 0.397 & 0.374 & 1.00 & 0.386 & 0.497 & 0.410 & 1.00 & 0.088 & 0.167 & 0.100 & 1.00 \\
Max Entropy & 0.304 & 0.398 & 0.376 & 0.95 & \textbf{0.390} & 0.501 & \textbf{0.414} & 0.99 & 0.088 & 0.167 & 0.100 & 1.00 \\
Max Probability & 0.304 & 0.396 & 0.374 & 1.00 & 0.386 & 0.497 & 0.410 & 1.00 & 0.088 & 0.167 & 0.100 & 1.00 \\
Mean CPMI & 0.302 & 0.397 & 0.376 & 0.98 & 0.386 & 0.497 & 0.410 & 1.00 & 0.090 & 0.169 & 0.102 & 0.99 \\
Mean Entropy & 0.306 & 0.400 & 0.378 & 0.93 & 0.386 & 0.497 & 0.410 & 0.99 & 0.088 & 0.167 & 0.100 & 1.00 \\
Mean PMI & 0.310 & 0.399 & 0.382 & 0.96 & 0.386 & 0.497 & 0.410 & 1.00 & 0.088 & 0.167 & 0.100 & 1.00 \\
Mean Probability & 0.308 & 0.400 & 0.380 & 0.96 & 0.388 & 0.498 & 0.412 & 0.97 & \textbf{0.092} & \textbf{0.173} & \textbf{0.104} & 0.97 \\
Median Entropy & 0.308 & 0.398 & 0.378 & 0.98 & 0.386 & 0.497 & 0.410 & 1.00 & 0.088 & 0.167 & 0.100 & 1.00 \\
Median Probability & 0.304 & 0.397 & 0.376 & 0.94 & 0.386 & 0.497 & 0.410 & 1.00 & 0.090 & 0.169 & 0.102 & 1.00 \\
Min Entropy & 0.308 & 0.397 & 0.376 & 0.93 & 0.386 & 0.497 & 0.410 & 1.00 & 0.090 & 0.171 & \textbf{0.104} & 0.99 \\
Min Probability & 0.312 & 0.401 & 0.376 & 0.95 & 0.386 & 0.497 & 0.410 & 1.00 & 0.090 & 0.169 & 0.102 & 0.99 \\
NumSemSets & 0.304 & 0.396 & 0.374 & 1.00 & 0.386 & \textbf{0.502} & 0.412 & 0.95 & 0.088 & 0.167 & 0.100 & 1.00 \\
PTrue & 0.308 & 0.398 & 0.372 & 0.87 & 0.386 & 0.497 & 0.410 & 1.00 & 0.090 & 0.169 & 0.102 & 0.99 \\
Perplexity & 0.304 & 0.398 & 0.376 & 0.96 & 0.386 & 0.498 & 0.410 & 1.00 & 0.088 & 0.168 & 0.100 & 1.00 \\
RDE & 0.304 & 0.398 & 0.376 & 0.99 & 0.386 & 0.497 & 0.410 & 1.00 & 0.090 & 0.171 & 0.102 & 0.99 \\
RMD & 0.304 & 0.398 & 0.372 & 0.97 & 0.388 & 0.499 & 0.412 & 0.95 & 0.088 & 0.167 & 0.100 & 1.00 \\
RenyiNeg & 0.302 & 0.396 & 0.374 & 1.00 & \textbf{0.390} & 0.500 & \textbf{0.414} & 0.97 & 0.088 & 0.167 & 0.100 & 1.00 \\
SAR & 0.310 & 0.404 & 0.380 & 0.97 & 0.386 & 0.500 & 0.412 & 0.90 & 0.088 & 0.167 & 0.100 & 1.00 \\
Semantic Entropy & 0.304 & 0.398 & 0.374 & 1.00 & 0.386 & 0.499 & 0.412 & 0.93 & 0.088 & 0.169 & 0.102 & 1.00 \\
SentenceSAR & 0.308 & 0.403 & 0.376 & 0.89 & 0.384 & 0.498 & 0.410 & 0.90 & 0.088 & 0.167 & 0.100 & 1.00 \\

\bottomrule
\end{tabular}
}
    \caption{Detailed QA performance results for uncertainty methods on one-hop datasets. `InAcc' denotes In-Accuracy, and `EM' stands for Exact Match. Higher values indicate better performance. Bold values highlight the best results. Standard deviations for InAcc, EM, and F1 are $\approx 0.02 \pm 0.002$ for HotPotQA and 2Wiki and $\approx 0.01 \pm 0.001$ for Musique, calculated using bootstrapping.
    }
    \label{tab:ue_result_multi_hop}
\end{table*}

\begin{table*}[t]
    \centering
    \small
    \resizebox{\textwidth}{!}{
\begin{tabular}{lccccc|ccccc|ccccc}
\toprule
\multirow{2}{*}{{Method}} & \multicolumn{5}{c|}{NQ} & \multicolumn{5}{c|}{SQUAD} & \multicolumn{5}{c}{TQA} \\
 & EM & F1 & Acc & LLMC & RC & EM & F1 & Acc & LLMC & RC & EM & F1 & Acc & LLMC & RC \\
\midrule
No Context & 0.386 & 0.495 & 0.446 & 1.0 & 0.00 & 0.156 & 0.249 & 0.176 & 1.0 & 0.00 & 0.592 & 0.675 & 0.636 & 1.0 & 0.00 \\
All Context & 0.388 & 0.506 & 0.496 & 1.0 & 1.00 & 0.252 & 0.389 & 0.312 & 1.0 & 1.00 & 0.522 & 0.636 & 0.610 & 1.0 & 1.00 \\
\midrule
AdaptiveRAG & 0.388 & 0.505 & 0.496 & 1.0 & 0.98 & 0.238 & 0.366 & 0.286 & 1.0 & 0.97 & 0.564 & 0.656 & 0.628 & 0.5 & 0.54 \\

DRAGIN & 0.396 & 0.510 & 0.480 & 4.5 & 2.24 & 0.244 & 0.371 & 0.298 & 4.3 & 2.14 & 0.584 & 0.691 & 0.666 & 4.1 & 2.06 \\
FLARE & 0.358 & 0.477 & 0.450 & 3.1 & 2.07 & 0.190 & 0.303 & 0.238 & 3.1 & 2.08 & 0.570 & 0.674 & 0.648 & 2.1 & 1.39 \\
FS-RAG & 0.348 & 0.483 & 0.428 & 2.7 & 2.70 & 0.226 & 0.361 & 0.286 & 2.8 & 2.78 & 0.540 & 0.640 & 0.632 & 2.5 & 2.47 \\
IRCoT & 0.392 & 0.502 & 0.478 & 2.7 & 2.70 & 0.210 & 0.341 & 0.268 & 2.7 & 2.68 & 0.526 & 0.634 & 0.608 & 2.7 & 2.74 \\

Rowen\textsubscript{CL} & 0.002 & 0.104 & 0.494 & 29.5 & 7.24 & 0.004 & 0.061 & 0.196 & 29.2 & 7.19 & 0.022 & 0.188 & 0.656 & 28.7 & 7.06 \\
Rowen\textsubscript{CM} & 0.002 & 0.104 & 0.494 & 29.5 & 7.27 & 0.004 & 0.061 & 0.196 & 29.2 & 7.20 & 0.022 & 0.188 & 0.656 & 28.7 & 7.12 \\
Rowen\textsubscript{Hybrid} & 0.002 & 0.104 & 0.494 & 55.0 & 7.27 & 0.004 & 0.061 & 0.196 & 54.3 & 7.15 & 0.022 & 0.189 & 0.656 & 53.4 & 6.93 \\
Seakr & 0.360 & 0.487 & 0.406 & 14.6 & 1.00 & 0.226 & 0.361 & 0.268 & 14.6 & 1.00 & 0.598 & 0.692 & 0.656 & 14.6 & 1.00 \\
\bottomrule
\end{tabular}
}
    \caption{Results of baselines for onehop datasets. LLMC refers to the average number of LLM calls per question, while RC indicates the average number of retrieval calls per question. For NQ the standard deviations of Acc, EM, and F1 are $\approx 0.022 \pm 0.001$ across all methods.  For SQUAD and Trivia the standard deviations of Acc, EM, and F1 are $\approx 0.018 \pm 0.006$ across all methods.
    Overall, the methods exhibit similar deviations, with Rowen showing the lowest deviation, typically $\leq 0.01$.
    }
    \label{tab:one_hop_baseline}
\end{table*}

\begin{table*}[t]
    \centering
    \small
    \resizebox{\textwidth}{!}{
\begin{tabular}{lccccc|ccccc|ccccc}
\toprule
\multirow{2}{*}{{Method}} & \multicolumn{5}{c|}{2Wiki} & \multicolumn{5}{c|}{HotPotQA} & \multicolumn{5}{c}{Musique} \\
 & EM & F1 & Acc & LLMC & RC & EM & F1 & Acc & LLMC & RC & EM & F1 & Acc & LLMC & RC \\
\midrule
No Context & 0.302 & 0.371 & 0.318 & 1.0 & 0.00 & 0.280 & 0.372 & 0.286 & 1.0 & 0.00 & 0.100 & 0.193 & 0.106 & 1.0 & 0.00 \\
All Context & 0.302 & 0.396 & 0.374 & 1.0 & 1.00 & 0.386 & 0.497 & 0.410 & 1.0 & 1.00 & 0.088 & 0.167 & 0.100 & 1.0 & 1.00 \\
\midrule
AdaptiveRAG & 0.384 & 0.471 & 0.454 & 2.6 & 2.64 & 0.396 & 0.499 & 0.414 & 2.3 & 2.34 & 0.122 & 0.216 & 0.140 & 3.6 & 3.63 \\
DRAGIN & 0.406 & 0.480 & 0.456 & 5.8 & 2.92 & 0.398 & 0.506 & 0.430 & 5.1 & 2.56 & 0.116 & 0.207 & 0.134 & 6.3 & 3.15 \\
FLARE & 0.358 & 0.451 & 0.424 & 3.9 & 2.85 & 0.298 & 0.391 & 0.372 & 5.1 & 4.07 & 0.076 & 0.161 & 0.090 & 4.1 & 3.10 \\
FS-RAG & 0.348 & 0.431 & 0.388 & 3.8 & 3.76 & 0.376 & 0.503 & 0.422 & 3.7 & 3.70 & 0.088 & 0.187 & 0.100 & 3.4 & 3.35 \\
IRCoT & 0.362 & 0.460 & 0.454 & 4.4 & 4.38 & 0.414 & 0.516 & 0.438 & 3.5 & 3.45 & 0.116 & 0.221 & 0.138 & 4.1 & 4.08 \\
Rowen\textsubscript{CL} & 0.002 & 0.083 & 0.444 & 32.9 & 7.87 & 0.002 & 0.084 & 0.354 & 31.9 & 7.67 & 0.002 & 0.034 & 0.104 & 42.1 & 9.52 \\
Rowen\textsubscript{CM} & 0.002 & 0.083 & 0.444 & 32.9 & 7.87 & 0.002 & 0.084 & 0.356 & 31.9 & 7.70 & 0.002 & 0.034 & 0.104 & 42.1 & 9.52 \\
Rowen\textsubscript{Hybrid} & 0.002 & 0.083 & 0.444 & 61.8 & 7.85 & 0.004 & 0.086 & 0.354 & 59.8 & 7.63 & 0.002 & 0.034 & 0.102 & 80.2 & 9.48 \\
Seakr & 0.382 & 0.460 & 0.398 & 12.3 & 2.44 & 0.400 & 0.523 & 0.424 & 9.9 & 1.76 & 0.112 & 0.215 & 0.118 & 12.3 & 2.40 \\
\bottomrule
\end{tabular}
}
    \caption{Results of baselines for multihop datasets. LLMC refers to the average number of LLM calls per question, while RC indicates the average number of retrieval calls per question. For 2Wiki and HotPotQA, the standard deviations of Acc, EM, and F1 are $\leq 0.022 \pm 0.001$ across all methods. For Musique, the standard deviations are $\leq 0.015 \pm 0.001$. Overall, the methods exhibit similar deviations, with Rowen showing the lowest deviation, typically $\leq 0.01$.
    }
    \label{tab:multi_hop_baseline}
\end{table*}


\begin{table*}[t]
    \centering
    \begin{tabular}{lcccccc|c}
    \toprule
     & NQ & SQUAD & TQA & 2Wiki & HotPot & Musique & Avg \\
    \midrule
    Mean CPMI & -2.02 & -7.44 & -4.38 & -2.98 & -5.76 & 0.78 & -3.63 \\
    Mean PMI & -1.45 & -8.21 & -4.69 & -4.19 & -5.37 & 3.20 & -3.45 \\
    RDE & -1.29 & -7.18 & -3.52 & -2.23 & -4.68 & 1.18 & -2.95 \\
    PTrue & -1.94 & -7.95 & -3.77 & -2.03 & -5.07 & 3.53 & -2.87 \\
    EigValLaplacian & -3.28 & -7.01 & -3.44 & -2.29 & -3.32 & 2.35 & -2.83 \\
    Min Probability & -1.83 & -5.26 & -3.52 & -2.98 & -3.22 & 1.96 & -2.48 \\
    RenyiNeg & -1.04 & -4.23 & -6.12 & 0.21 & -3.38 & 2.40 & -2.03 \\
    NumSemSets & -0.96 & -5.90 & -3.39 & -0.21 & -3.20 & 1.60 & -2.01 \\
    FisherRao & -0.80 & -5.26 & -3.44 & -2.75 & -3.12 & 3.60 & -1.96 \\
    Min Entropy & -2.22 & -4.23 & -3.08 & -0.85 & -1.17 & 0.38 & -1.86 \\
    Median Entropy & -0.89 & -3.97 & -3.81 & -1.48 & -3.80 & 3.60 & -1.72 \\
    Median Probability & -1.13 & -2.44 & -4.60 & -0.11 & -3.41 & 1.96 & -1.62 \\
    Hybrid & -1.83 & -4.71 & -4.28 & -0.32 & -5.12 & 7.06 & -1.53 \\
    Mean Probability & -0.16 & -2.39 & -4.22 & -0.42 & -2.43 & 1.54 & -1.35 \\
    Max Entropy & -1.90 & -2.56 & -4.80 & -0.43 & -3.00 & 6.40 & -1.05 \\
    CCP & 0.32 & -2.18 & -6.77 & -0.64 & -2.44 & 6.00 & -0.95 \\
    Max Probability & -0.64 & -2.95 & -4.41 & -0.64 & -2.24 & 5.20 & -0.95 \\
    DegMat & 0.56 & -2.69 & -3.91 & -1.57 & -2.15 & 4.80 & -0.83 \\
    Lex-Similarity & -2.50 & -3.77 & -3.41 & -0.85 & -2.34 & 8.00 & -0.81 \\
    Eccentricity & 0.00 & -2.31 & -2.63 & -2.60 & -2.71 & 5.60 & -0.78 \\
    SAR & -0.24 & -2.31 & -2.87 & -1.05 & -3.11 & 5.60 & -0.66 \\
    SentenceSAR & 0.32 & -2.42 & -3.35 & -0.85 & -2.15 & 5.20 & -0.54 \\
    Semantic Entropy & -0.48 & -1.77 & -2.69 & 0.43 & -1.94 & 3.53 & -0.49 \\
    RMD & 0.08 & -7.56 & -1.76 & -0.11 & -0.49 & 7.60 & -0.37 \\
    Perplexity & 0.24 & -2.66 & -4.27 & 0.85 & -1.56 & 5.60 & -0.30 \\
    Mean Entropy & -0.48 & -1.40 & -3.94 & 0.74 & -1.76 & 7.20 & 0.06 \\
    MD & 0.00 & -2.56 & -2.74 & 0.65 & 0.00 & 9.60 & 0.82 \\
    \bottomrule
    \end{tabular}
    \caption{Average QA performance differences after transfer (in percentage) for each dataset. Negative values indicate a loss in In-Accuracy compared to in-domain testing, while positive values represent an In-Accuracy gain.}
    \label{tab:qa_transfer}
\end{table*}

\begin{table*}[ht!]
    \centering
    \small
    \resizebox{\textwidth}{!}{
\begin{tabular}{lll}
\toprule
\multicolumn{1}{c}{\textbf{\begin{tabular}[c]{@{}c@{}}Method\\ acronym\end{tabular}}} & \multicolumn{1}{c}{\textbf{\begin{tabular}[c]{@{}c@{}}Method \\ full name\end{tabular}}}           & \multicolumn{1}{c}{\textbf{Short description}}                                                                                                                                                                                                                                                                                                                                                                                                              \\ \midrule
\multicolumn{3}{c}{\textbf{logit based}}                                                                                

\\ \midrule

\vspace{1.5ex}

\begin{tabular}[c]{@{}l@{}}FisherRao \\ \cite{DBLP:conf/emnlp/DarrinPC23}\end{tabular}                                                                              & \begin{tabular}[c]{@{}l@{}}Fisher-Rao\\ distance\end{tabular}                                      & \begin{tabular}[c]{@{}l@{}}FisherRao is a distance on the Riemannian space formed by the parametric distributions, using the Fisher\\ information matrix as its metric. It computes the geodesic distance between two discrete distributions.

\vspace{1.1ex}

\end{tabular}                                                                                                                                                                                                  \\
\begin{tabular}[c]{@{}l@{}}Max Entropy \\ \cite{fomicheva2020unsupervised}\end{tabular}                                                                           & \begin{tabular}[c]{@{}l@{}}Maximum Token \\ Entropy\end{tabular}                                   & The maximum entropy of all tokens in the generated  sequence.                                                                  \\
Max Probability                                                                       & \begin{tabular}[c]{@{}l@{}}Maximum Sequence\\ Probability\end{tabular}                             & The score leverages the probability of the most likely sequence generation.                                                                                                                                                                                                                                                                                                                                                                                 \\
\begin{tabular}[c]{@{}l@{}}Mean CPMI \\ \cite{DBLP:conf/emnlp/PoelCM22}\end{tabular}                                                                             & \begin{tabular}[c]{@{}l@{}}Mean conditional\\ pointwise mutual\\ information\end{tabular}          & \begin{tabular}[c]{@{}l@{}}Extension of the PMI method by considering only those marginal probabilities for which the entropy of the\\ conditional distribution is above certain threshold.\end{tabular}                                                                                                                                                                                                                                                    \\
\begin{tabular}[c]{@{}l@{}}Mean Entropy \\ \cite{fomicheva2020unsupervised}\end{tabular}                                                                          & \begin{tabular}[c]{@{}l@{}}Mean Token\\ Entropy\end{tabular}                                       & The average entropy of each individual token in the generated sequence.                                                                       \vspace{1.1ex} 

\\
\begin{tabular}[c]{@{}l@{}}Mean PMI \\ \cite{takayama2019relevant}\end{tabular}                                                                              & \begin{tabular}[c]{@{}l@{}}Mean pointwise\\ mutual information\end{tabular}                        & \begin{tabular}[c]{@{}l@{}}PMI compares the probability of two events (the  question and the generated answer)  occurring together to what\\ this probability would be if the events were independent.\end{tabular}                                                                                                                                                                                                                                         \\
Mean Probability                                                                      & \begin{tabular}[c]{@{}l@{}}Mean Sequence\\ Probability\end{tabular}                                & The total uncertainty is measured via average sequence probability.                                                                                                                                                                                                                                                                                                                                                                                         \\
\begin{tabular}[c]{@{}l@{}}Median Entropy \\ \cite{fomicheva2020unsupervised}\end{tabular}                                                                        & \begin{tabular}[c]{@{}l@{}}Median Token\\ Entropy\end{tabular}                                     & The median entropy of all tokens in the generated sequence.                                                                                                                                                                                                                                                                                                                                                                                                 \\
Median Probability                                                                    & \begin{tabular}[c]{@{}l@{}}Median Sequence\\ Probability\end{tabular}                              & The total uncertainty is measured via  median sequence probability.                                                                                                                                                                                                                                                                                                                                                                                         \\
\begin{tabular}[c]{@{}l@{}}Min Entropy \\ \cite{fomicheva2020unsupervised}\end{tabular}                                                                           & \begin{tabular}[c]{@{}l@{}}Minimum Token\\ Entropy\end{tabular}                                    & The minimum entropy of all tokens in the generated sequence.                                                                                                                                                                                                                                                                                                                                                                                                \\
Min Probability                                                                       & \begin{tabular}[c]{@{}l@{}}Minimum Sequence\\ Probability\end{tabular}                             & The score leverages the probability of the least likely sequence generation.                                                                                                                                                                                                                                                                                                                                                                                \\
\begin{tabular}[c]{@{}l@{}}Perplexity \\ \cite{fomicheva2020unsupervised}\end{tabular}                                                                            & Perplexity                                                                                         & The score computes the average negative log probability of generated tokens, which is further exponentiated.                          
\vspace{1.1ex} 

\\
\begin{tabular}[c]{@{}l@{}}PTrue \\ \cite{kadavath2022language}\end{tabular}                                                                                & probability P(true)                                                                                & \begin{tabular}[c]{@{}l@{}}The method measures the uncertainty of the claim  by asking the LLM itself whether the generated claim is true or\\ not.  The confidence is the probability of thefirst generated token y1 being equal to “True”.\end{tabular}                                                            \vspace{1.1ex}                                                                                                                                         \\
\begin{tabular}[c]{@{}l@{}}RenyiNeg  \\ \cite{DBLP:conf/emnlp/DarrinPC23}\end{tabular}                                                                               & Rényi negentropy                                                                                   & The score computes alpha-Renyi-divergence between the sample and the uniform distributions.                                                                                 \vspace{1.1ex}                                                                                                \\
\begin{tabular}[c]{@{}l@{}}SAR\\  \cite{duan2023shifting}\end{tabular}                                                                                    & \begin{tabular}[c]{@{}l@{}}Shifting Attention\\ to more Relevant\end{tabular}                      & \begin{tabular}[c]{@{}l@{}}SAR corrects generative inequalities by reviewing the relevance of each token and emphasizing uncertainty\\ quantification attention to those more relevant components. The relevance is measured by calculating similarity\\ of sentence before and after removing the certain token.\end{tabular}                                                                                                                              \\
\begin{tabular}[c]{@{}l@{}}SentenceSAR\\  \cite{duan2023shifting}\end{tabular}                                                                           & \begin{tabular}[c]{@{}l@{}}Shifting Attention\\ to more Relevant \\ at Sentence level\end{tabular} & SAR measured at sentence-level.                                                                                                                                                                                                                                                                                                                                                                                                                             \\ \midrule
\multicolumn{3}{c}{\textbf{consistency based}}                                                                                                                                                                                                                                                                                                                                                                                                                                                                                                                                                                                                     \\ \midrule
\vspace{1.5ex} 

\begin{tabular}[c]{@{}l@{}}CCP  \\ \cite{DBLP:conf/acl/FadeevaRSPLMTKP24}\end{tabular}                                                                                  & \begin{tabular}[c]
{@{}l@{}}Claim-Conditioned\\ Probability\end{tabular}                                                & \begin{tabular}[c]{@{}l@{}}The method aggregates token-level uncertainties into a claim-level score, it removes the impact of uncertainty \\ about what claim to generate on the current step and what surface form to use. 

\vspace{1.1ex}

\end{tabular} \\

\begin{tabular}[c]{@{}l@{}}DegMat  \\ \cite{lin2023generating}\end{tabular}                                                                                 & \begin{tabular}[c]{@{}l@{}}Degree \\ matrix
\end{tabular}                                           & Using the Degree matrix a new uncertainty measure could be found that reflects the average pairwise distance.        \vspace{1.1ex} 

\\
\begin{tabular}[c]{@{}l@{}}Eccentricity  \\ \cite{lin2023generating}\end{tabular}                                                                           & Eccentricity                                                                                       & \begin{tabular}[c]{@{}l@{}}The smallest k eigenvectors of Laplacian Graph are used as the proxy for the models’ embeddings. Then, we could\\ use the average offset from the average embedding as the uncertainty measure.\end{tabular}                                                                          \vspace{1.1ex}                                                                                                                                             \\
\begin{tabular}[c]{@{}l@{}}EigValLaplacian \\ \cite{lin2023generating}\end{tabular}                                                                       & \begin{tabular}[c]{@{}l@{}}Sum of  eigenvalues of \\ the graph Laplacian\end{tabular}              & \begin{tabular}[c]{@{}l@{}}The score uses pairwise similarities between the sampled answers to the questions to form the symmetric weighted\\ adjacency matrix (degree matrix). This matrix is further used to create the graph Laplacian. The sum of Eigenvalues\\ of the Graph Laplacian are used as a measure of uncertainty.\end{tabular}                       \vspace{1.1ex} 

\\
\begin{tabular}[c]{@{}l@{}}Lex-Similarity \\ \cite{fomicheva2020unsupervised}\end{tabular}                                                                       & \begin{tabular}[c]{@{}l@{}}Lexical \\ similarity\end{tabular}                                      & The score computes how similar two words or phrases are in terms of their meaning.                                                                                                                                                                                                                                                                                                                                                                          \\
\begin{tabular}[c]{@{}l@{}}NumSemSets   \\ \cite{lin2023generating}\end{tabular}                                                                            & \begin{tabular}[c]{@{}l@{}}Number of semantic\\ sets\end{tabular}                                  & \begin{tabular}[c]{@{}l@{}}The number of semantic sets initially equals the total number of generated answers K. If two answers are\\ semantically similar, they are put into one cluster. A higher number of semantic sets corresponds to an increased\\ level of uncertainty, as it suggests a higher number of diverse semantic interpretations for the answer.
 \vspace{1.1ex} 
 
\end{tabular}                                                                             \\ 

\begin{tabular}[c]{@{}l@{}}Semantic Entropy \\  \cite{DBLP:conf/iclr/KuhnGF23}\end{tabular}                                                                      & \begin{tabular}[c]{@{}l@{}}Semantic\\ Entropy\end{tabular}                            & \begin{tabular}[c]{@{}l@{}}The method aims to deal with the generated sequences that have similar meaning while  having different  probabilities \\ according to the model. The idea is to cluster generated sequences into several semantically  homogeneous \\ clusters with a bi-directional entailment algorithm and average the sequence  probabilities within the clusters.\end{tabular} \\

\midrule
\multicolumn{3}{c}{\textbf{internal-based}}                                                                                                                                                                                                                                                                                                                                                                                                                                                                                                                                                                                                           \\ \midrule
\vspace{1.5ex}

\begin{tabular}[c]{@{}l@{}}MD \\ \cite{DBLP:conf/nips/LeeLLS18}\end{tabular}                                                                                   & 
 
\begin{tabular}[c]{@{}l@{}}Mahalanobis\\ distance\end{tabular}                                     & \begin{tabular}[c]{@{}l@{}}In this paper, the authors propose a simple yet effective method for detecting any abnormal samples, which is\\ applicable to any pre-trained softmax neural classifier. They obtain the class conditional Gaussian distributions\\ with respect to (low- and upper-level) features of the deep modelsunder Gaussian discriminant analysis, which\\ result in a confidence score based on the Mahalanobis distance.\end{tabular} 
\vspace{1.1ex} 
 
\\
\begin{tabular}[c]{@{}l@{}}RDE \\ \cite{DBLP:conf/acl/YooKJK22}\end{tabular}                                                                                   & \begin{tabular}[c]{@{}l@{}}Robust density\\ estimation\end{tabular}                                & \begin{tabular}[c]{@{}l@{}}The method improves over MD by reducing the dimensionality of the last hidden state of the decoder averaged\\ over all generated tokens via PCA decomposition.  Additionally, computing  of the covariance matrix for each\\ individual class is done by using the Minimum Covariance Determinant estimation. The uncertainty score is\\ computed as  the MD in the space of reduced dimensionality.
\vspace{1.1ex} 
 
\end{tabular}                \\
\begin{tabular}[c]{@{}l@{}}RMD  \\ \cite{DBLP:conf/iclr/0006LZKSLL23}\end{tabular}                                                                                   & \begin{tabular}[c]{@{}l@{}}Relative Mahalanobis\\ distance\end{tabular}                            & \begin{tabular}[c]{@{}l@{}}The MD distance score is adjusted by subtracting from it the other  MD score computed for some\\ large general purpose dataset covering many domains.
\vspace{1.1ex} 
 
 \end{tabular}                                                                                                                                                                                                                                                               \\  \midrule
\multicolumn{3}{c}{\textbf{blended approach}}                                                                   \\ \midrule

Hybrid                                                                                     & 
 Hybrid
                                   &
\begin{tabular}[c]{@{}l@{}} Our hybrid approach that uses all uncertainty features defined in the table.\end{tabular} 
\vspace{1.1ex} 
 
\\

\bottomrule

\end{tabular}

}
   
    \caption{Description of the uncertainty estimation methods used in the paper. The methods are grouped by their categories: logit based, consistency-based, internal-based and hybrid.
     \label{tab:ue_desc}
 }

\end{table*}\begin{table*}[t]
    \centering
    \begin{tabular}{lccccc}
    \toprule
    Method & EigValLaplacian & Lex-Similarity & Max Entropy & Mean Entropy & SAR \\
    \midrule
    EigValLaplacian & 1.00 & 0.03 & 0.81 & 0.00 & 0.00 \\
    Lex-Similarity & 0.03 & 1.00 & 0.38 & 0.42 & 0.95 \\
    Max Entropy & 0.81 & 0.38 & 1.00 & 0.00 & 0.08 \\
    Mean Entropy & 0.00 & 0.42 & 0.00 & 1.00 & 0.86 \\
    SAR & 0.00 & 0.95 & 0.08 & 0.86 & 1.00 \\
    \bottomrule
    \end{tabular}
    \caption{In-Accuracy P-Value, Friedman Test Results:
Test Statistic: 29.580
P-value: 0.00001}
    \label{tab:qa_ood_test}
\end{table*}

\begin{table*}[t]
    \centering
    \begin{tabular}{lccccc}
    \toprule
    Method & EigValLaplacian & Lex-Similarity & Max Entropy & Mean Entropy & SAR \\
    \midrule
    EigValLaplacian & 1.00 & 0.00 & 0.08 & 0.00 & 0.05 \\
    Lex-Similarity & 0.00 & 1.00 & 0.81 & 0.94 & 0.88 \\
    Max Entropy & 0.08 & 0.81 & 1.00 & 0.33 & 1.00 \\
    Mean Entropy & 0.00 & 0.94 & 0.33 & 1.00 & 0.42 \\
    SAR & 0.05 & 0.88 & 1.00 & 0.42 & 1.00 \\
    \bottomrule
    \end{tabular}
    \caption{Accuracy P-Value, Friedman Test Results:
Test Statistic: 22.847
P-value: 0.00014}
    \label{tab:self_knowledge_ood_test}
\end{table*}


\section{Performance Analysis Across Datasets}
\begin{figure*}[htbp]
    \centering
    \includegraphics[width=0.99\textwidth]{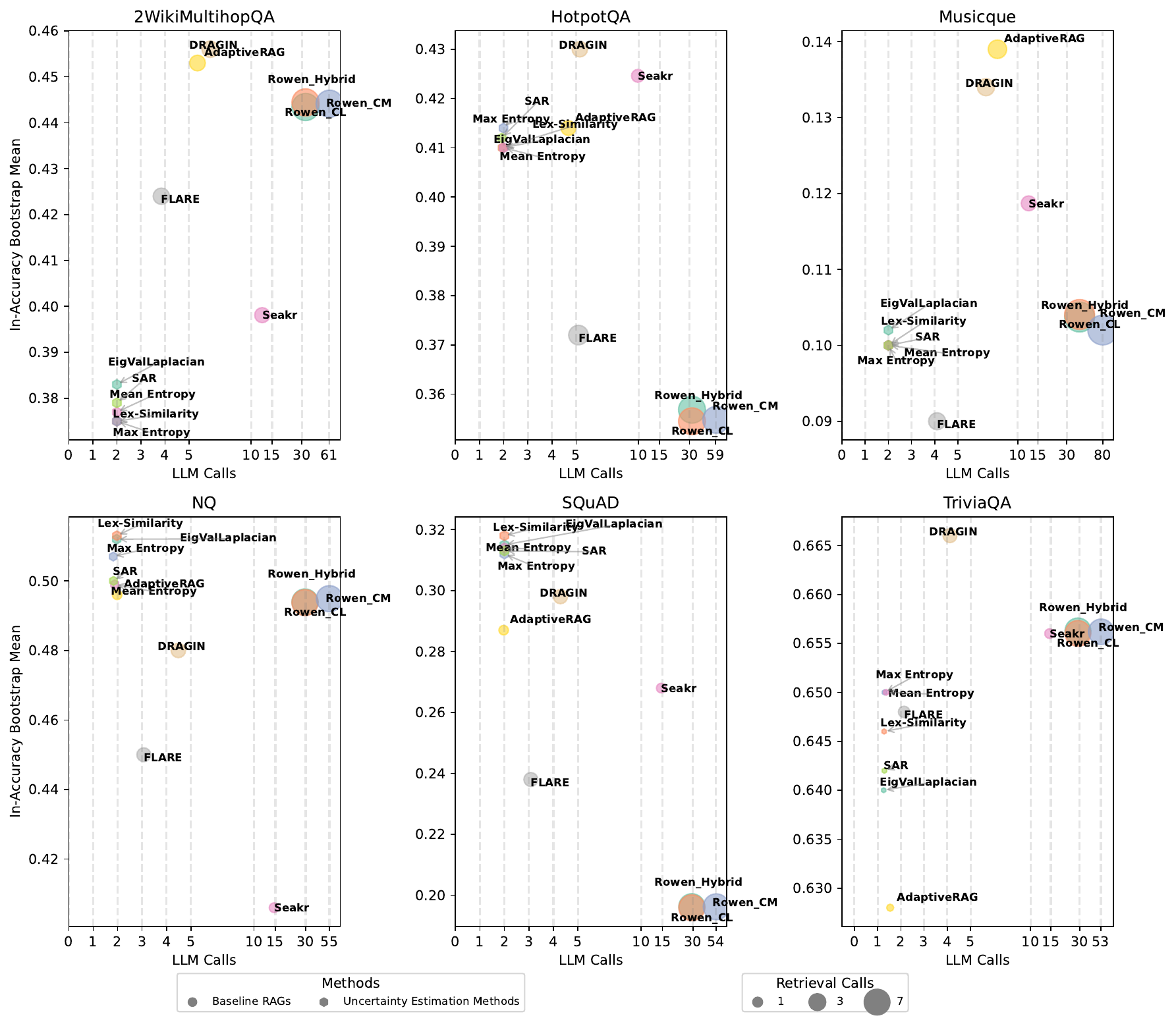}
    \caption{Performance comparison showing the relationship between LLM calls and Bootstrap Mean In-Accuracy. The size of each point indicates the number of retrieval calls required by each method.}
    \label{fig:inacc_vs_llmcall}
\end{figure*}

The scatter plot visualizes the performance comparison of various Retrieval-Augmented Generation (RAG) methods for all studied datasets, Figure~\ref{fig:inacc_vs_llmcall}.
The x-axis represents the number of LLM calls, while the y-axis shows the Bootstrap Mean In-Accuracy. Circle sizes in the visualization correspond to the number of retrieval calls required by each method. 

\clearpage
\section{Performance Analysis Across OOD Datasets}\label{sec:appendix_ood_heatmap}

\begin{figure*}[ht!]
\centering
\begin{minipage}[b]{.4\textwidth}
\includegraphics[width=\linewidth]{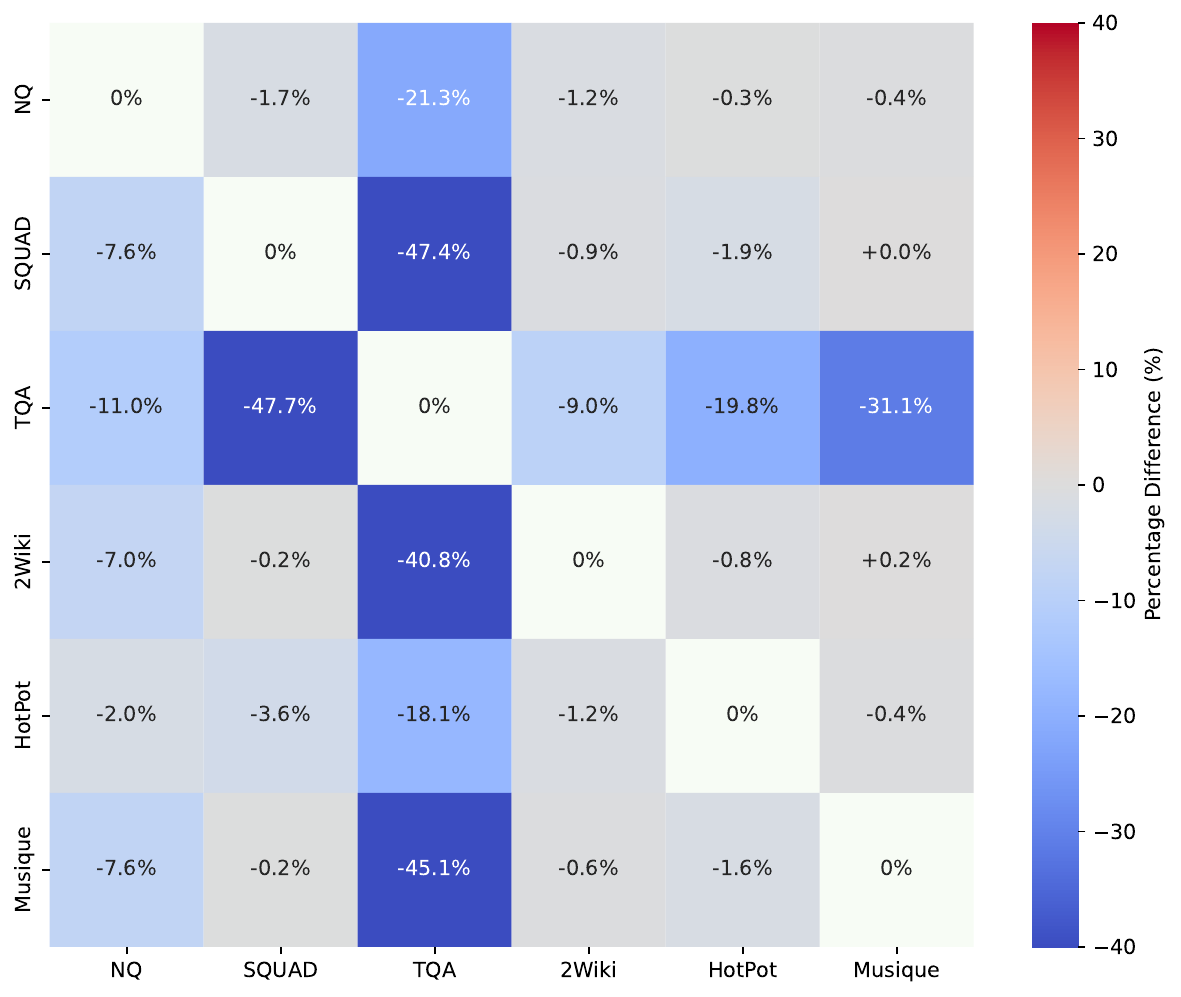}
\end{minipage}\qquad
\begin{minipage}[b]{.4\textwidth}
\includegraphics[width=\linewidth]{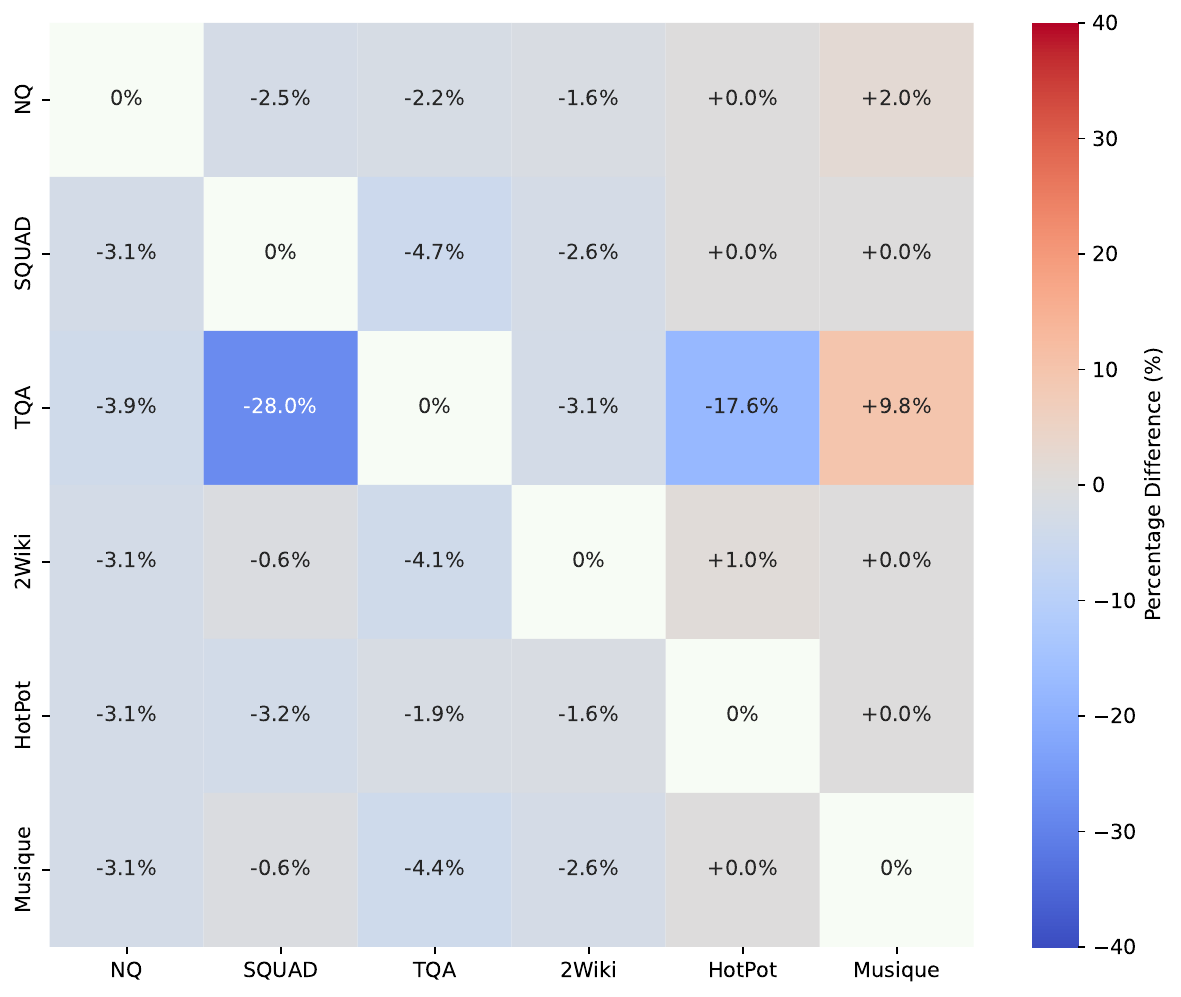}
\end{minipage}
\caption{Heatmap of improvement/decrease of the Accuracy and In-Accuracy scores on the OOD setup for the EigValLaplacian method.}
    \label{fig:eig_acc}
\end{figure*}

\begin{figure*}[ht!]
\centering
\begin{minipage}[b]{.4\textwidth}
\includegraphics[width=\linewidth]{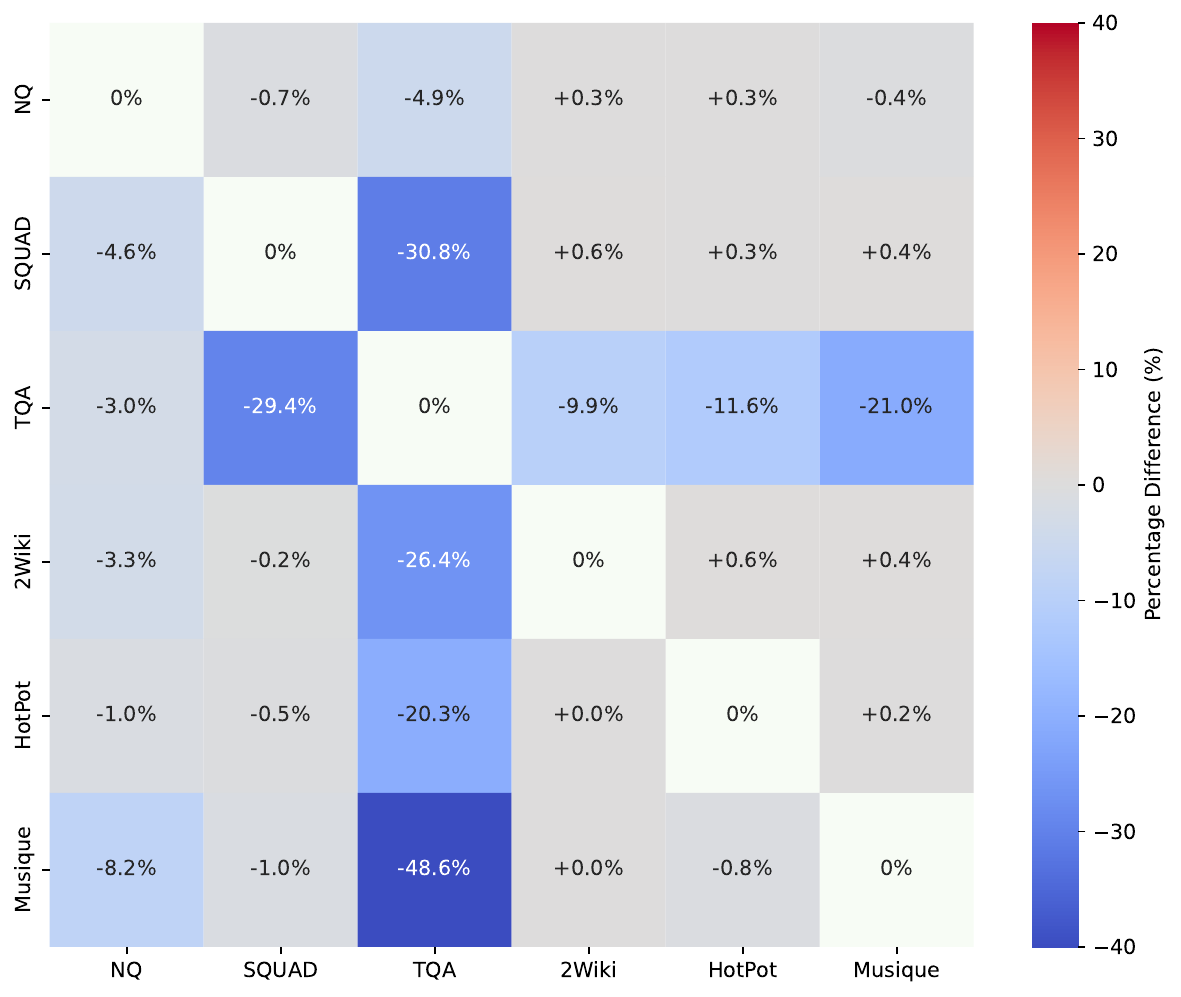}
\end{minipage}\qquad
\begin{minipage}[b]{.4\textwidth}
\includegraphics[width=\linewidth]{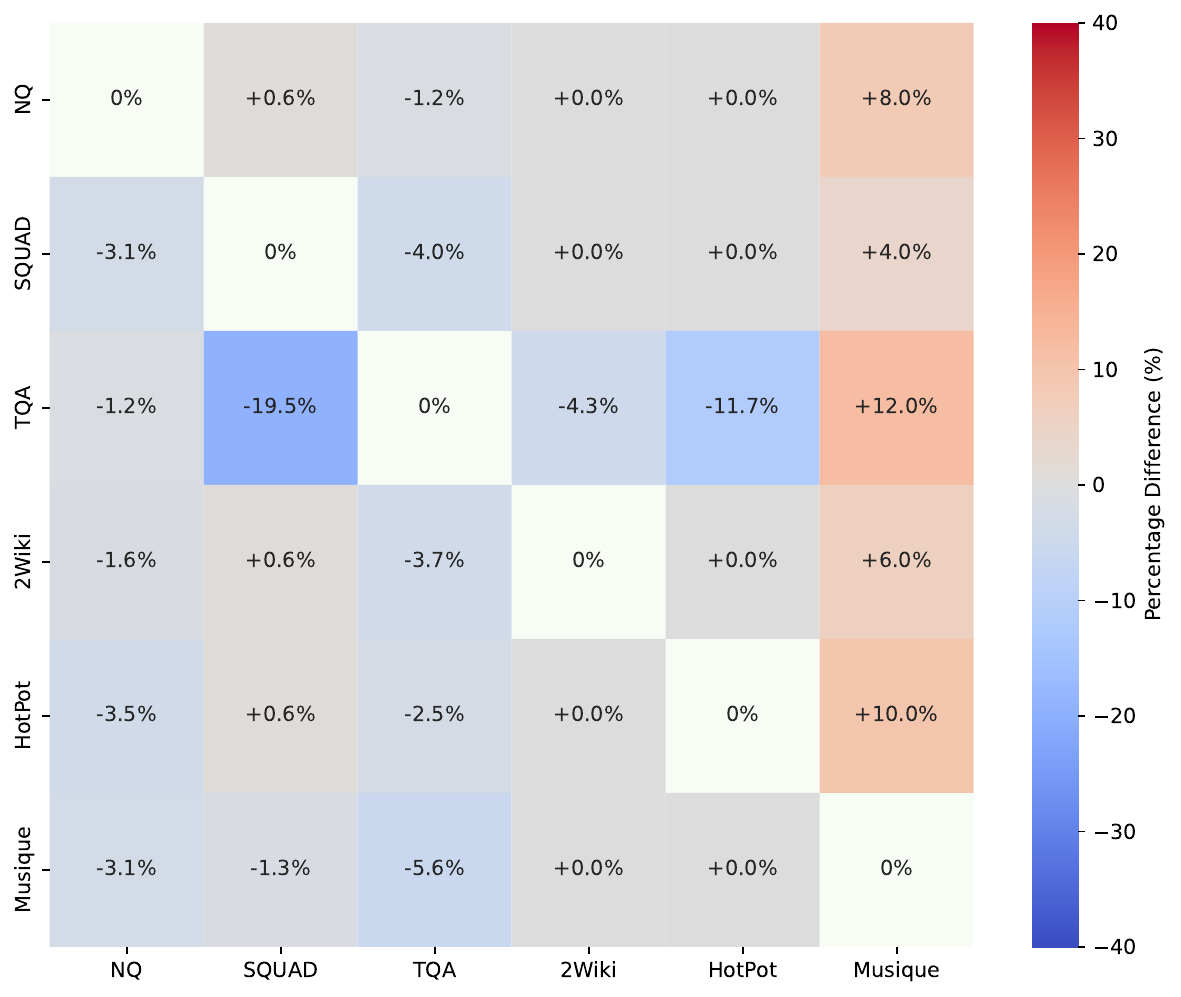}
\end{minipage}
\caption{Heatmap of improvement/decrease of the Accuracy and In-Accuracy scores on the OOD setup for the Lex-Similarity method.}
    \label{fig:lex_acc}
\end{figure*}

\begin{figure*}[ht!]
\centering
\begin{minipage}[b]{.4\textwidth}
\includegraphics[width=\linewidth]{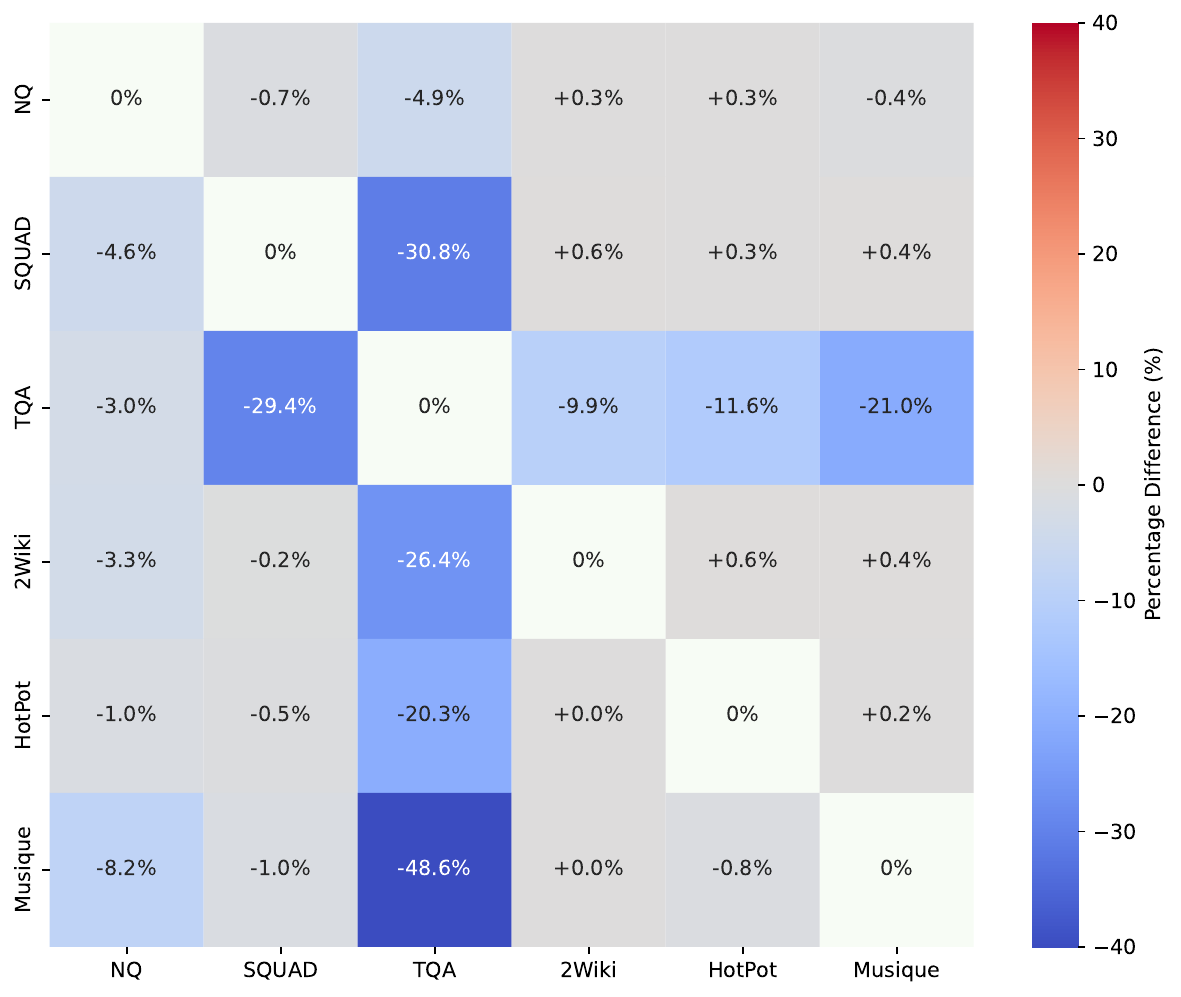}
\end{minipage}\qquad
\begin{minipage}[b]{.4\textwidth}
\includegraphics[width=\linewidth]{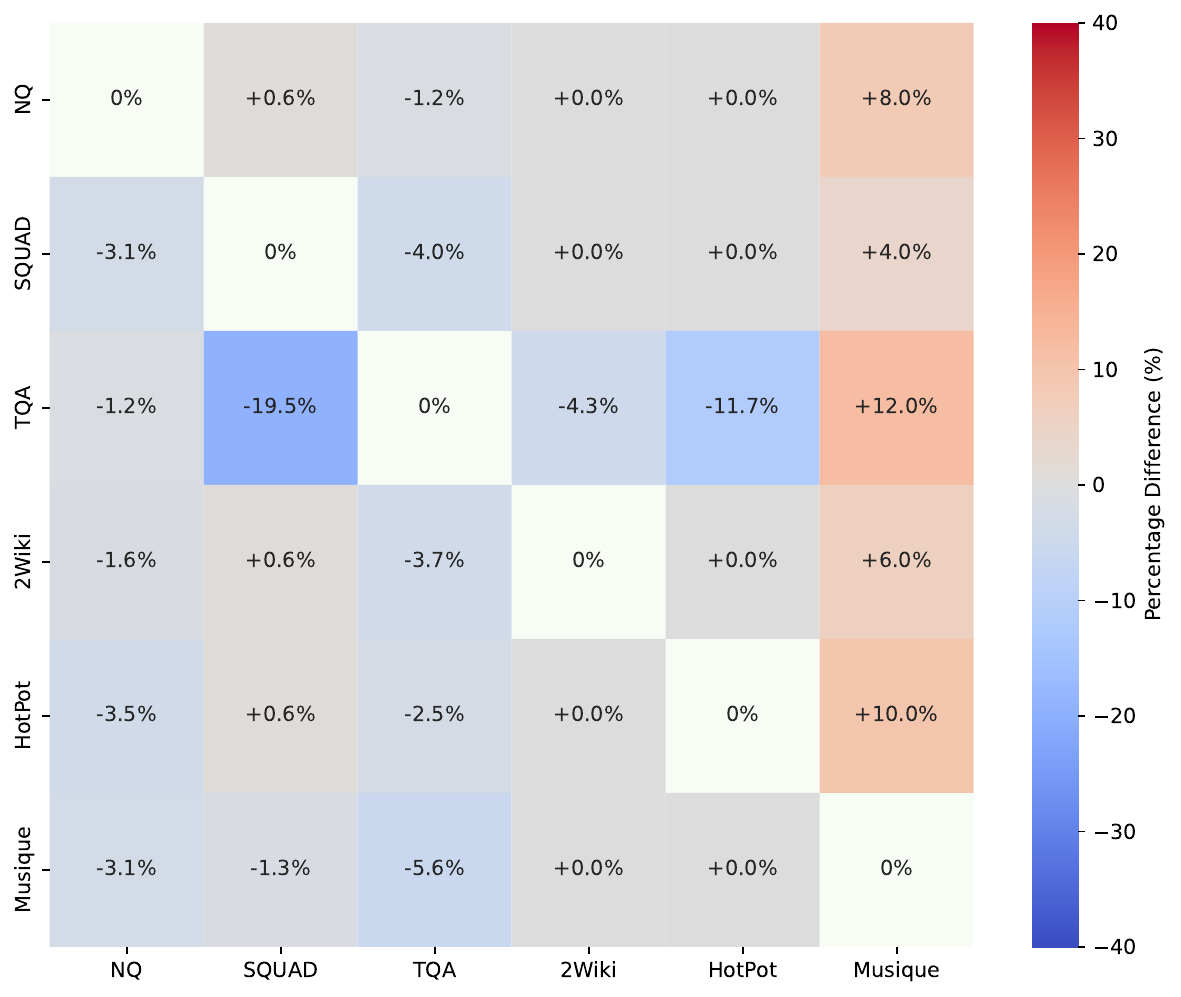}
\end{minipage}
\caption{Heatmap of improvement/decrease of the Accuracy and In-Accuracy scores on the OOD setup for the MaxEntropy method.}
    \label{fig:max_acc}
\end{figure*}

\begin{figure*}[ht!]
\centering
\begin{minipage}[b]{.4\textwidth}
\includegraphics[width=\linewidth]{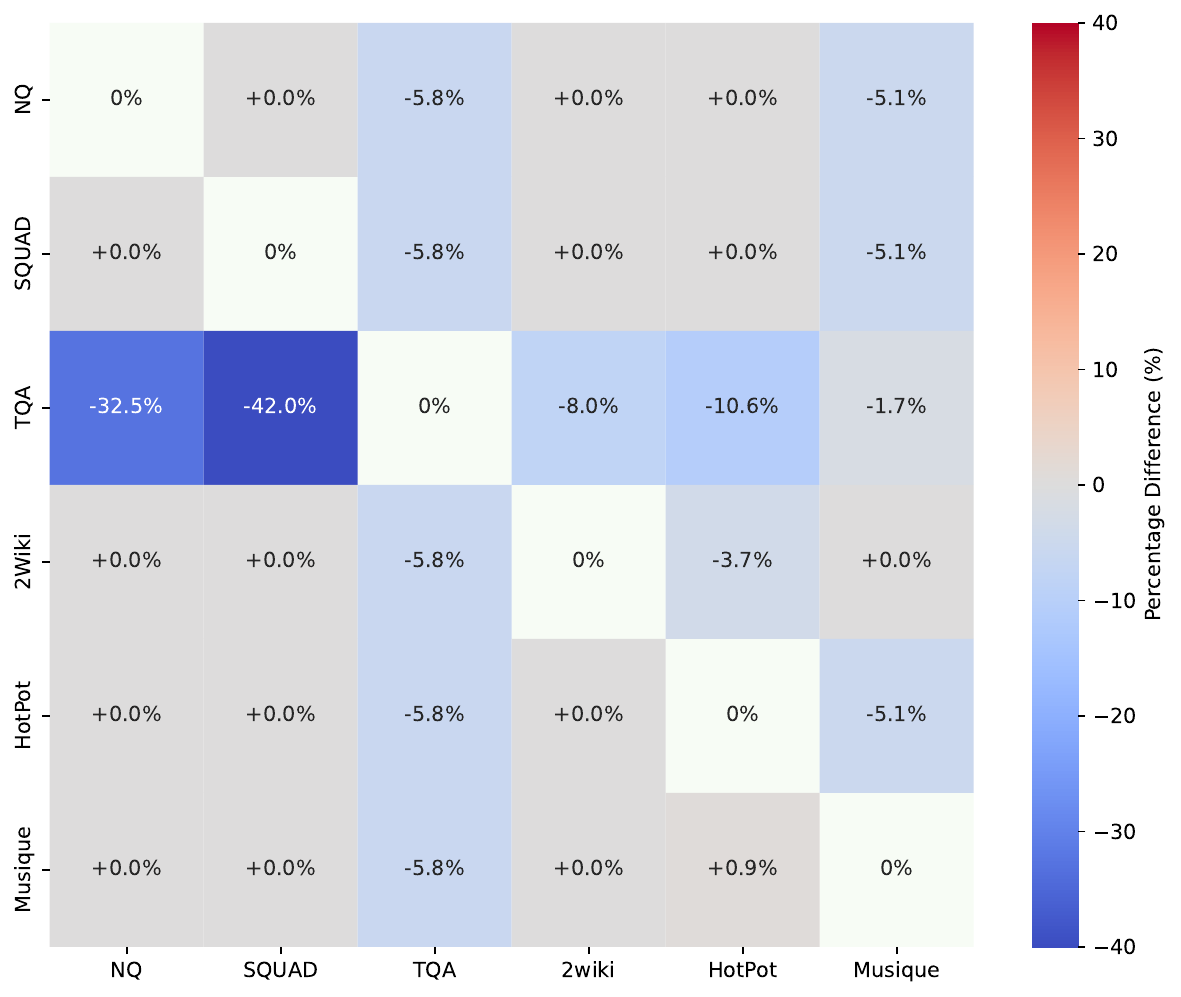}
\end{minipage}\qquad
\begin{minipage}[b]{.4\textwidth}
\includegraphics[width=\linewidth]{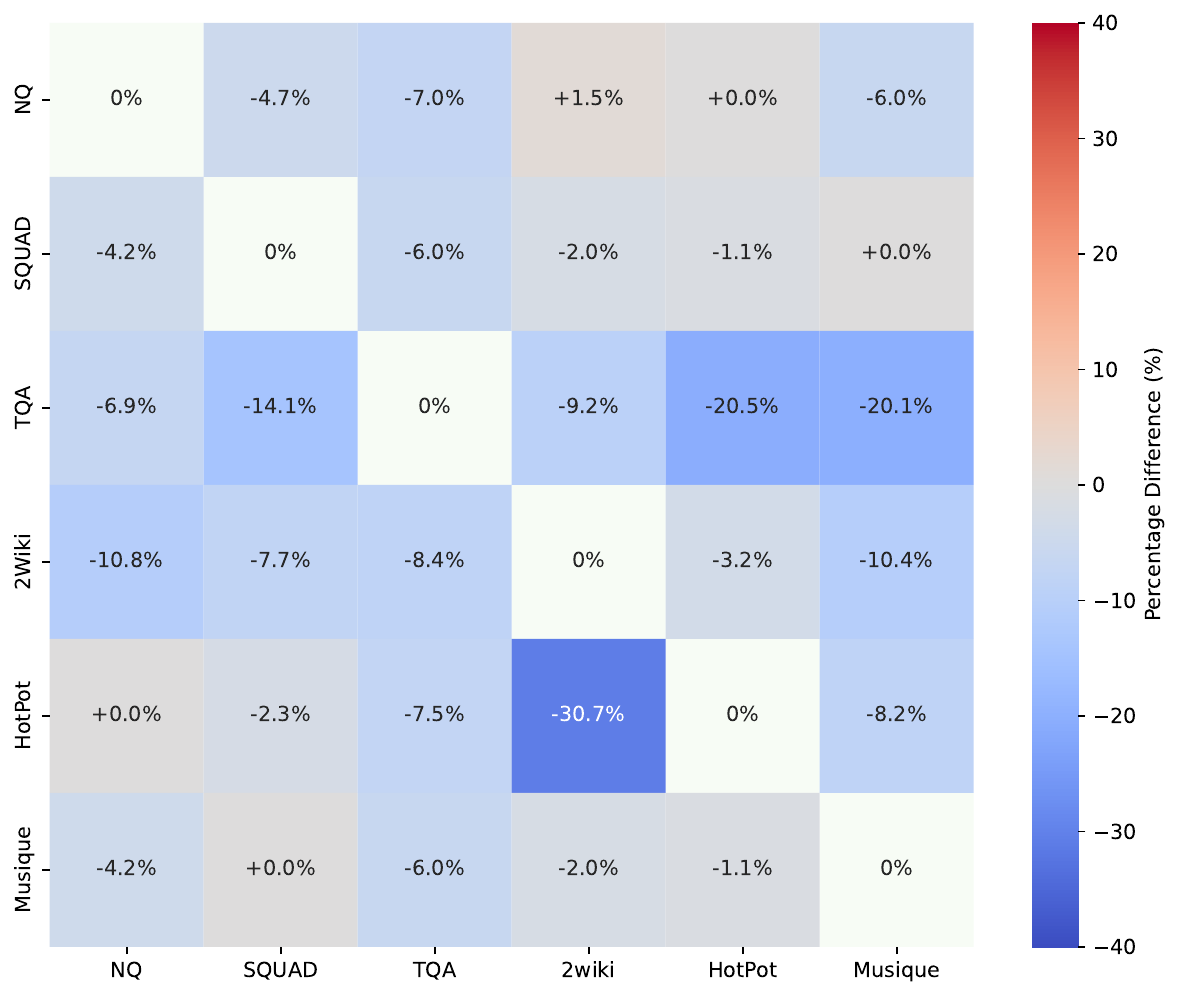}
\end{minipage}
\caption{Heatmap of improvement/decrease of the In-Accuracy scores on the OOD setup for the SeaKR and DRAGIN methods}
    \label{fig:eig_seakr_dragin_acc}
\end{figure*}

\begin{figure*}[ht!]
\centering
\begin{minipage}[b]{.4\textwidth}
\includegraphics[width=\linewidth]{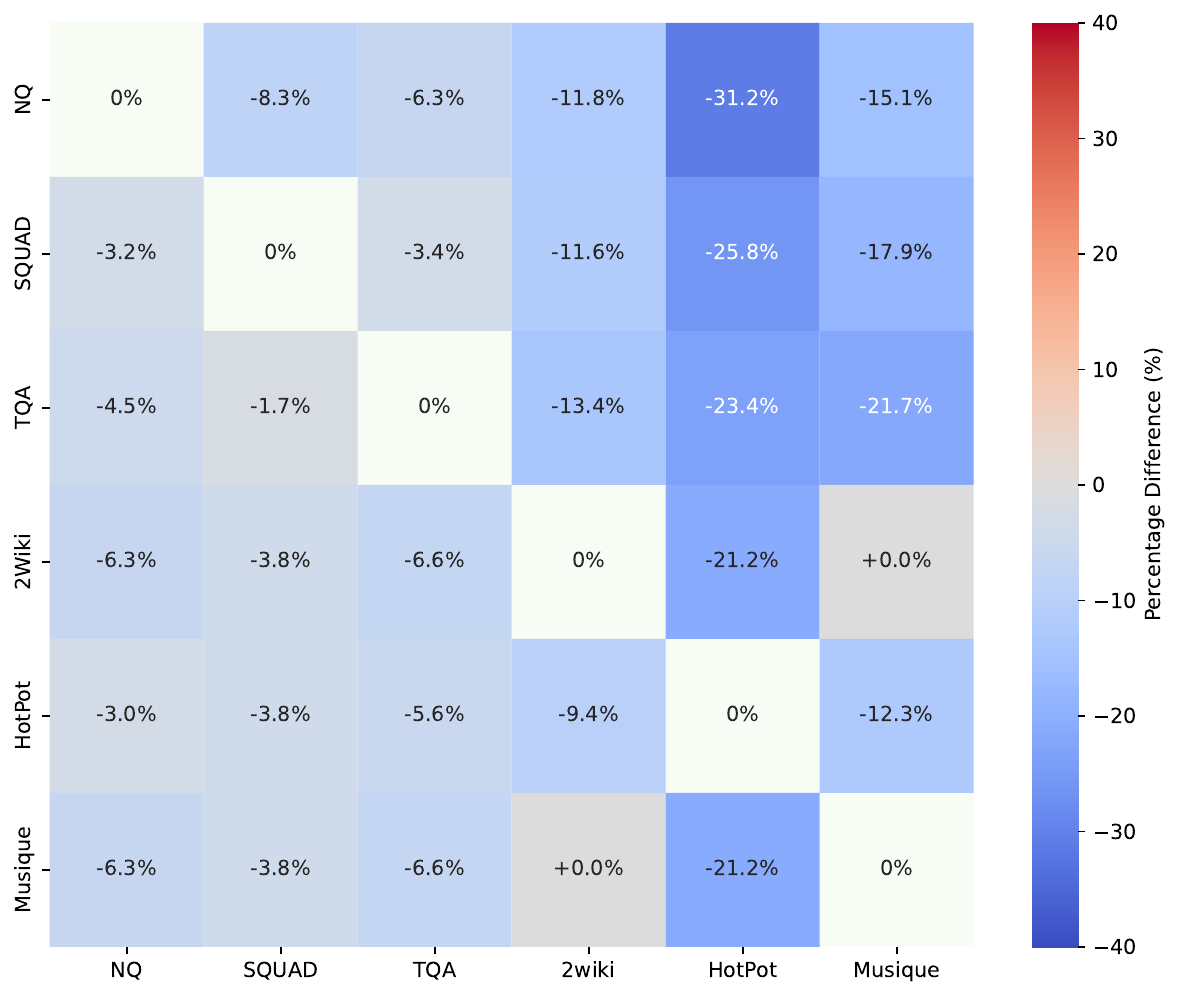}
\end{minipage}\qquad
\begin{minipage}[b]{.4\textwidth}
\includegraphics[width=\linewidth]{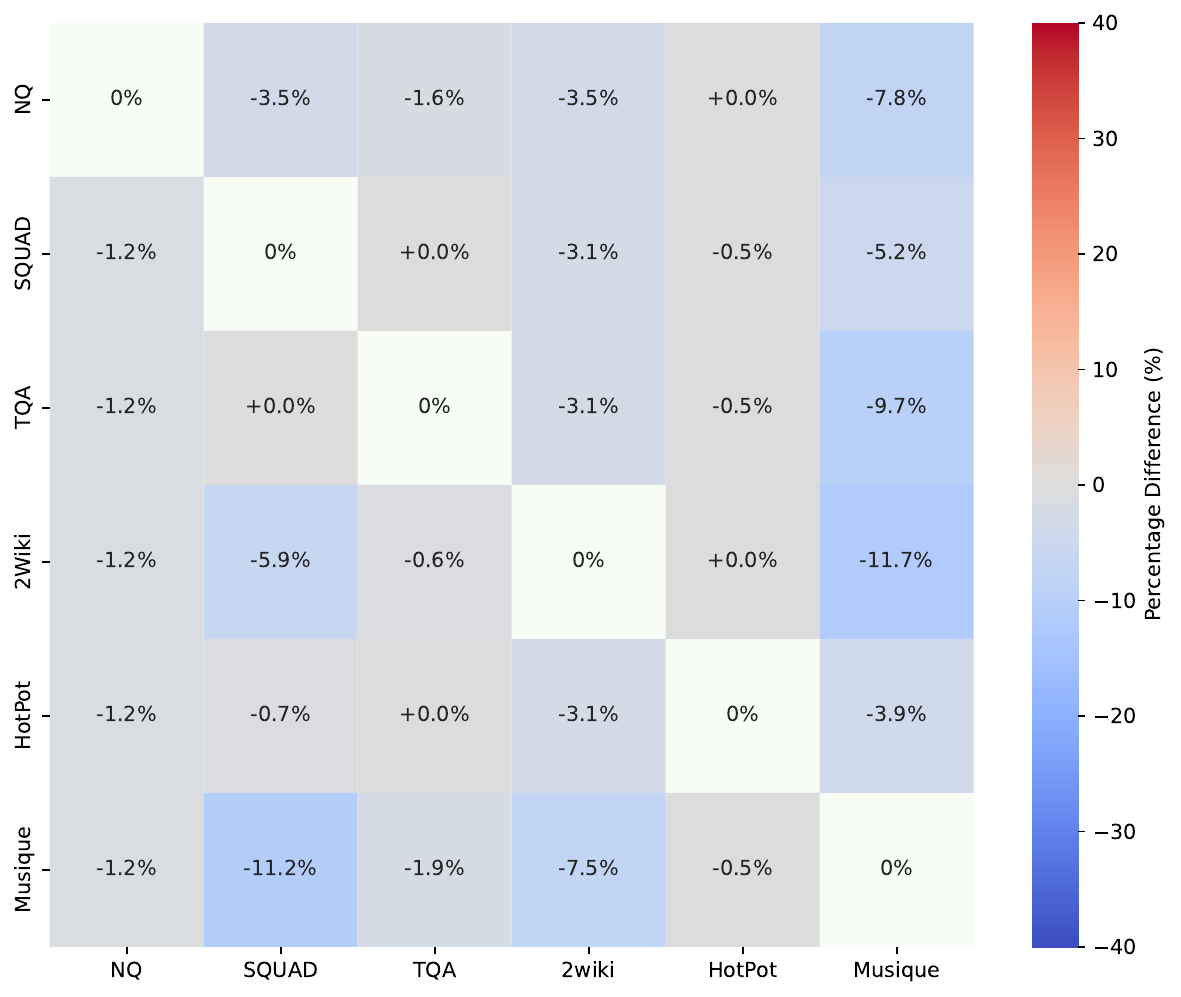}
\end{minipage}
\caption{Heatmap of improvement/decrease of the In-Accuracy scores on the OOD setup for the FLARE and AdaptiveRAG methods. AdaptiveRAG shows the most stable performance in OOD.}
    \label{fig:eig_flare_adarag_acc}
\end{figure*}

\clearpage
\section{Feature Importance Analysis for Hybrid Method of Uncertainty Estimation} \label{appx:fi_hybrid_ue}

\begin{figure*}[ht!]
    \centering
    \includegraphics[width=0.9\textwidth]{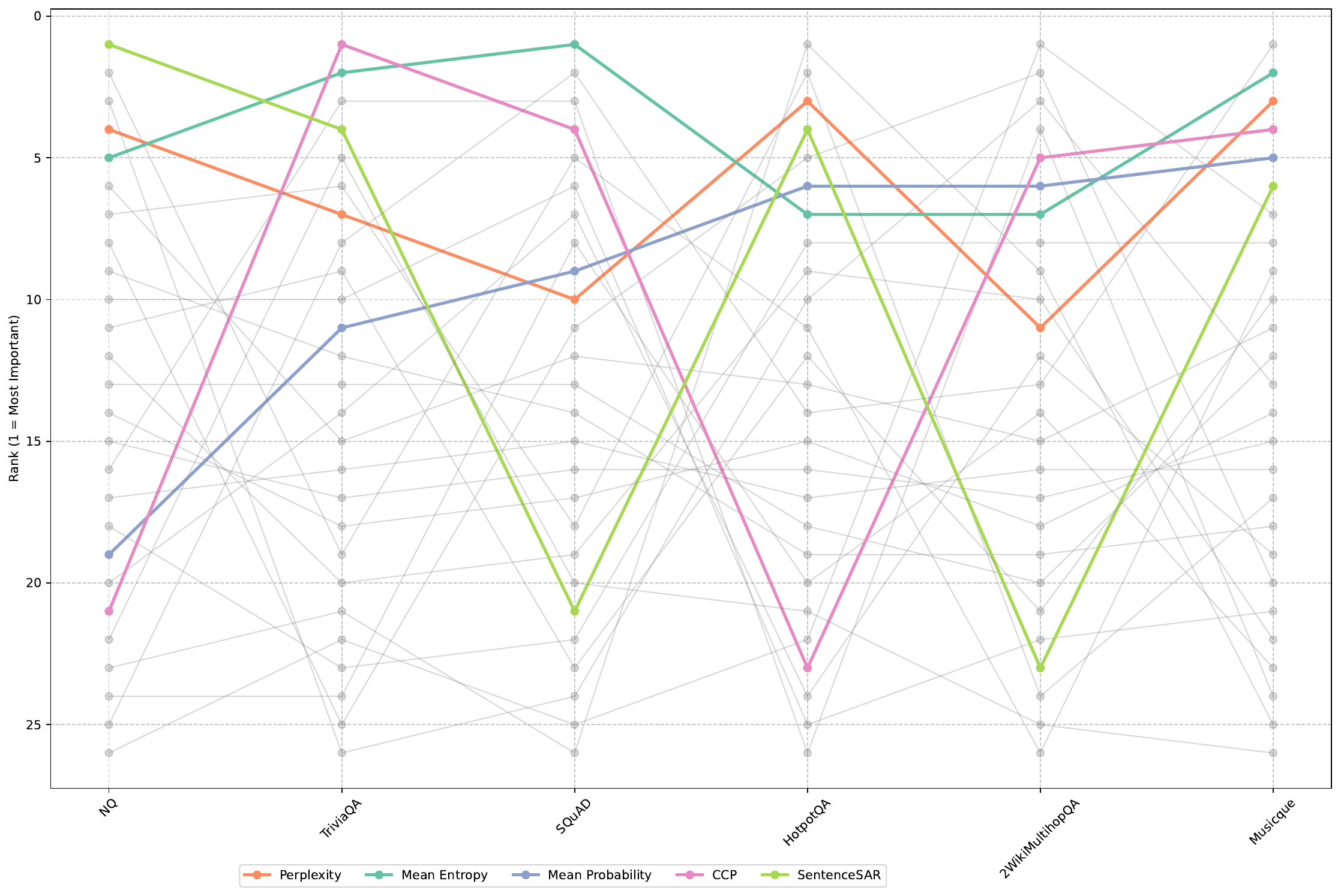}
    \caption{Top-5 UE methods as a features for hybrid method across datasets.}
    \label{fig:FI_rankings}
\end{figure*}

\begin{figure*}[ht!]
    \centering
    \includegraphics[width=0.99\textwidth]{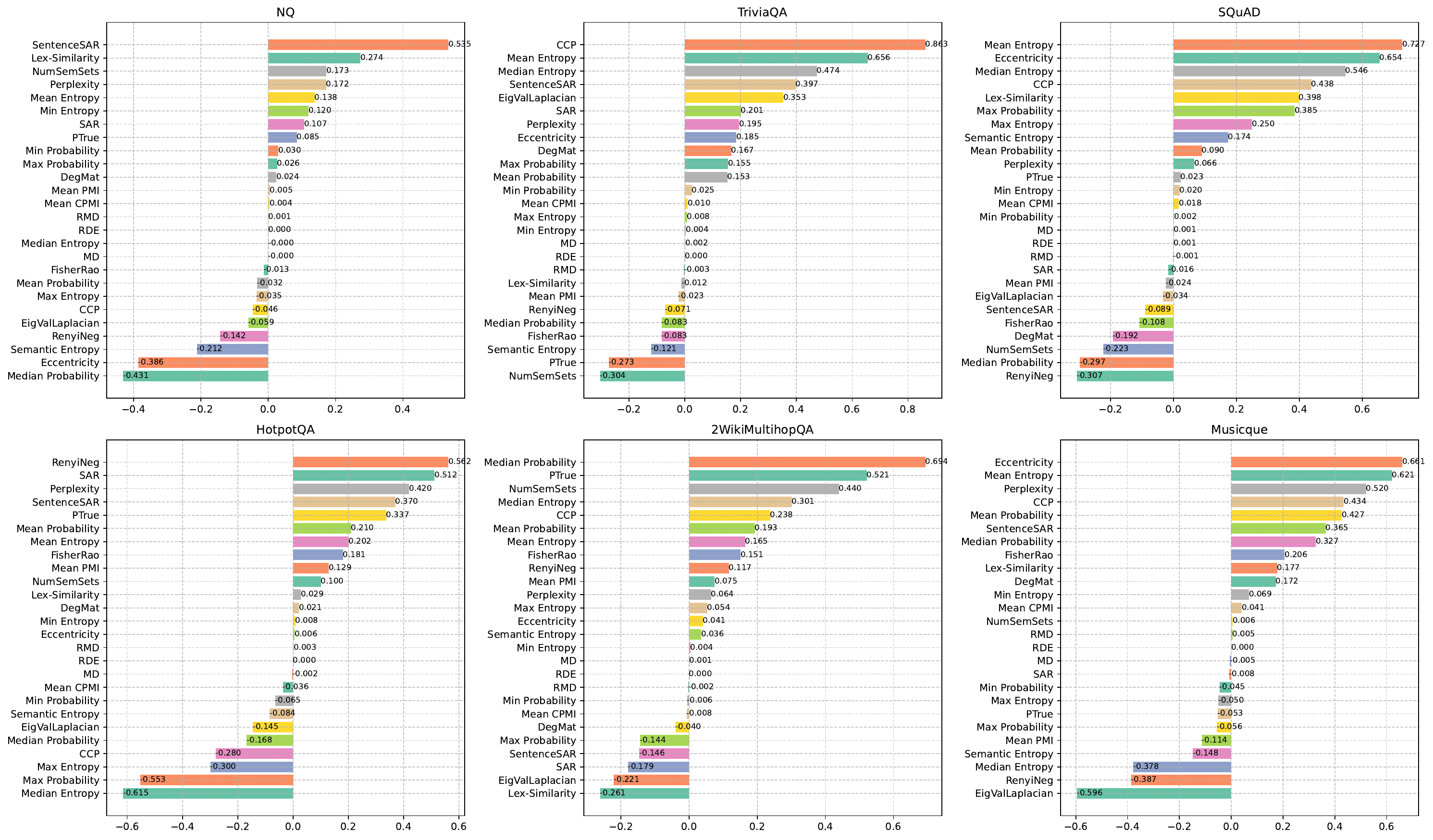}
    \caption{Feature Importance for each dataset for Hybrid method.}
    \label{fig:FI_barcharts}
\end{figure*}

This section provides a figure~\ref{fig:FI_rankings} to represent ranks importance of different uncertainty estimation methods as a feature in a hybrid method. In addition, Figure~\ref{fig:FI_barcharts} represents feature importance estimation in the form of a bar chart for each dataset.

\begin{figure}[ht!]
    \centering
    \includegraphics[width=0.99\linewidth]{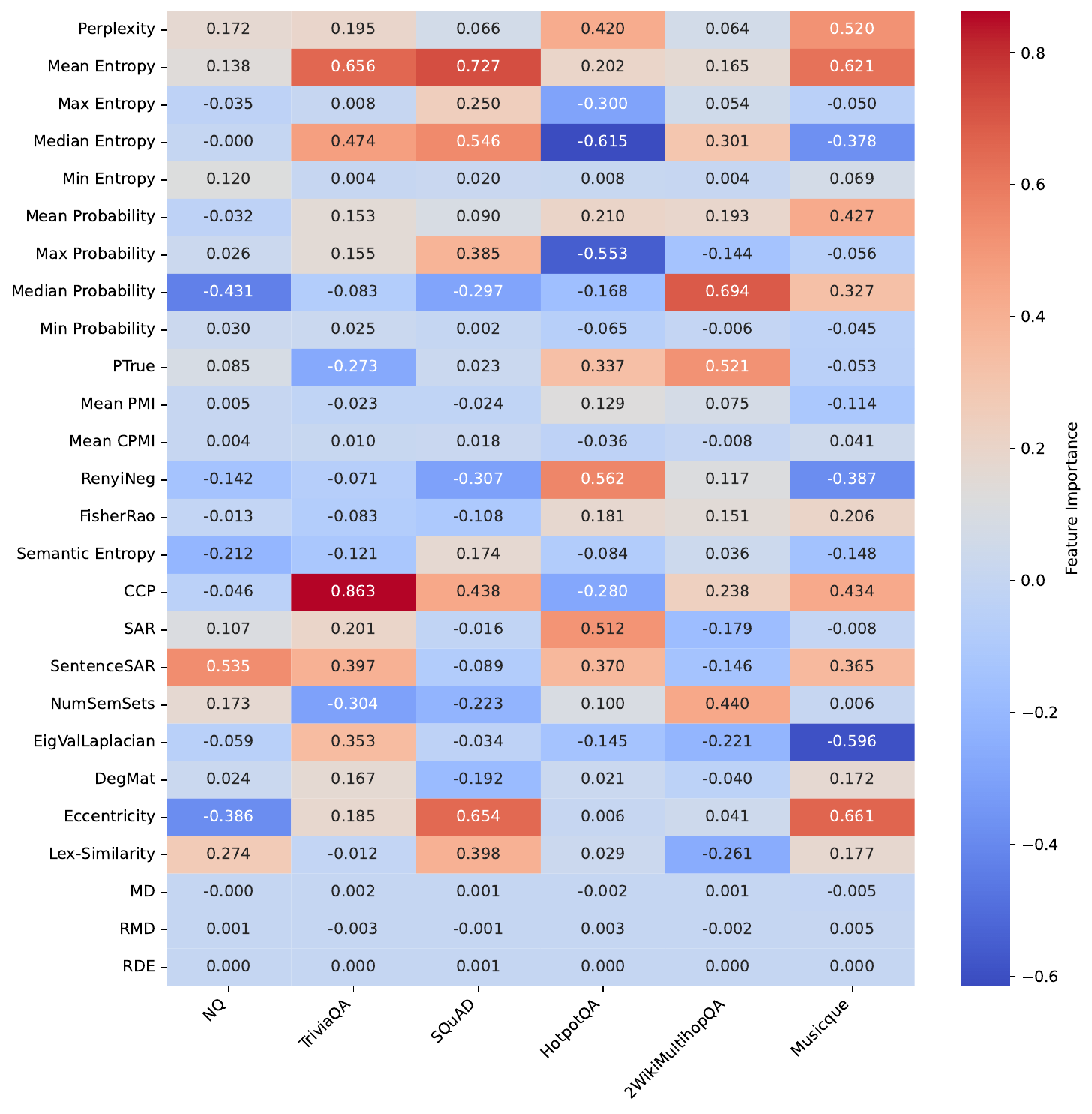}
    \caption{Feature Importance across datasets for Hybrid method. Different Uncertainty Estimation methods showed different performance on different datasets.}
    \label{fig:fi_heatmap}
\end{figure}

\clearpage
\section{Technical Details} \label{sec:appendix_technical}

For all experiments, we use the LLaMA 3.1-8b-instruct model with its default generation parameters. In our baseline methods, we strictly adhere to their original procedures, including prompting, parameter settings, and other configurations. For testing uncertainty estimation methods, we follow the protocol of AdaptiveRAG~\cite{DBLP:conf/naacl/JeongBCHP24}, using the same prompt and few-shot examples. For the Rowen Consistency Model evaluation we use Qwen~2.5-72B-Instruct~\cite{qwen2, qwen2.5} as the verification model instead of the original Qwen-Max-0428 due to the API usage limitations.

For all our methods we use the same retriever, a term-based sparse retrieval model known as BM25~\cite{DBLP:conf/trec/RobertsonWJHG94} and the same version of Elasticsearch 7.17.9\footnote{\url{https://www.elastic.co/guide/en/elasticsearch/reference/7.17/release-notes-7.17.9.html}}, following previous studies~\cite{DBLP:conf/acl/SuTA0024, DBLP:journals/corr/abs-2406-19215}. For the external document corpus, we use the Wikipedia corpus preprocessed by~\citet{DBLP:conf/emnlp/KarpukhinOMLWEC20}.

For all uncertainty methods, we compute scores on both the training and test sets with LM-Polygraph~\cite{DBLP:conf/emnlp/FadeevaVTVPFVGP23} with MIT License. Using the training set scores, we fit multiple classifiers, including Threshold, Logistic Regression, Decision Tree, KNN, and MLP. The performance of the best classifier is reported based on downstream metrics, with a further analysis of classifier stability provided in Section~\ref{sec:classifier_sens}. 

For classifiers, we employed scikit-learn library \cite{sklearn_api} and the following configurations:

\begin{itemize}
    \item Logistic Regression with default hyperparameters.
    \item Threshold Classifier is optimized by finding the best threshold for In-Accuracy over a log-scaled grid of size 200, spanning the minimum to maximum training uncertainty values.
    \item Decision Tree with a maximum depth of 3.
    \item K-Nearest Neighbors (KNN) using 15 neighbors.
    \item Multi-Layer Perceptron (MLP) configured with 2 hidden layers, each of size 64.
\end{itemize}

All hyperparameters remained fixed across all runs to ensure consistency.

Standard deviation is calculated via bootstrap sampling using 1000 rounds.

For trainable uncertainty methods, such as Mahalanobis Distance, we split the training data into two equal parts: one part is used to learn the parameters of the uncertainty method, while the other is used to train the classifier. For Relative Mahalanobis Distance, we utilize C4 as the source of additional relative data for training the parameters.

Our evaluation was conducted on an NVIDIA A100 GPU. The total runtime was approximately 6 hours for SeaKR, 18 hours for both DRAGIN and FLARE, 36 hours for IRCoT, 2 hours for training AdaptiveRAG on IRCoT generations, and 10 hours for Rowen\textsubscript{CM}, Rowen\textsubscript{CL}, and Hybrid (with caching). In contrast, all uncertainty estimations at once required less than 1 hour, highlighting their computational efficiency and reduced CO\textsubscript{2} emissions.

\section{Methods} \label{sec:baselines_description}

\paragraph{IRCoT}~\cite{DBLP:conf/acl/TrivediBKS23} -- \textbf{I}nterleaves \textbf{R}etrieval in a \textbf{CoT} was one of the pioneering methods to work with multi-hop questions. The authors proposed a new approach that interleaves retrieval with steps in chain-of-thought (CoT) reasoning. At first, the authors rertrieve \textit{K} paragraphs relevant to the question \textit{Q} as a query. Next, there are two steps, namely, \texttt{reason} and \texttt{retrieve} that are made iteratively until the termination criterion is met. As the incontext examples, questions, answers, gold relevant contexts and the example of CoT for the question are shown. In the \texttt{reason} step, we show the model CoT reasoning generated so far and let it complete the rest. Although the model can generate multiple sentences in the CoT, only the first generated sentence is taken. If the CoT contains the phrase "answer is:" (which was shown in the context examples as a phrase after which the final answer is written, so that we fix the format of the answers) or the maximum number of iterations has been reached \footnote{set to 8 in the experiments}, the process is terminated. In the \texttt{retrieve} step the last generated sentence in the CoT is taken to retrieve more additional paragraphs that would be relevant to answer the questions. These newly retrieved paragraphs are added to the ones retrieved in the previous question \footnote{maximum amount of paragraphs is set to 15 to fit the model's context limit} as the context for the question.

\paragraph{Adaptive RAG}~\cite{DBLP:conf/naacl/JeongBCHP24} uses the complexity of the question for adaptive retrieval. Simple questions can be answered without retrieval at all while complex questions require a multistep approach with iterative usage of both LLMs and retrievers. While users often ask simple and straightforward questions, the strategy which is necessary for answering complex questions is largely inefficient for the simple queries. The authors proposed a balanced strategy by training a classifier that predicts one of the three outcomes: whether not to retrieve at all (class A), retrieve once (class B) and retrieve multiple times (class C, the authors use IRCoT~\cite{DBLP:conf/acl/TrivediBKS23}). The classifier based on the t5-large model is trained on the development parts of the six considered datasets. The authors ran questions for all three methods and labeled as the correct the most efficient one. The most efficient means that if the correct answer is obtained by all three classes, class A is returned as the true one. As the additional training data the authors used the inductive biases in datasets (this concept assumes that simple questions should be answered with one step retrieve, and complex questions with multistep retrieve). 

\paragraph{FLARE}~\cite{jiang-etal-2023-active} -- \textbf{F}orward-\textbf{L}ooking \textbf{A}ctive \textbf{Re}trieval augmented generation is a method designed to improve the performance of LLMs by selectively integrating external knowledge. The idea behind FLARE is to monitor the probabilities generated by the LLM during the generation of the answers. If the model generates a token with probability below threshold (i.e. the model is uncertain), FLARE intervenes by querying an external knowledge source, such as a search engine or structured knowledge base, to retrieve relevant information. Using this additional context, FLARE regenerates the response until the next uncertain token or ends the generation. This approach balances high-quality generation and high response speed.

\paragraph{DRAGIN}~\cite{DBLP:conf/acl/SuTA0024} -- \textbf{D}ynamic \textbf{R}etrieval \textbf{A}ugmented \textbf{G}eneration based on the \textbf{I}nformation \textbf{N}eeds of LLMs follows a similar approach to FLARE by monitoring the model's token probabilities during generation. If LLM produces tokens with low likelihood, indicating uncertainty or knowledge gaps, DRAGIN triggers a retrieval process. For better identification of uncertainty tokens, DRAGIN filters out all stopwords. \footnote{\url{https://spacy.io/usage/linguistic-features}}. This paper also introduces an additional step: reformulating the query with keywords before retrieving information. These reformulated keywords are based on the model's internal attention weights and reasoning, allowing the system to determine what information is necessary and target relevant external knowledge sources more effectively. By incorporating new knowledge and ensuring the relevance of the retrieved information, DRAGIN improves the coherence of the final response. This approach reduces the risk of retrieving irrelevant documents and optimizes the model's reasoning process, especially in situations where queries may be ambiguous or incomplete.

\paragraph{Rowen}~\cite{DBLP:journals/corr/abs-2402-10612-rowen} -- \textbf{R}etrieve \textbf{O}nly \textbf{W}h\textbf{en} It Needs method presents a novel approach to reducing hallucinations in LLMs. This method uses an adaptive retrieval mechanism to improve the accuracy of LLM output. The method intelligently determines when to use external knowledge sources, based on a language and model evaluation. 

The Rowen Consistency Language~(Rowen~CL) component of Rowen involves generating semantically equivalent perturbations of the input query across English and Chinese languages. This includes asking the model to produce variations of the same question and then comparing the consistency of the responses generated in different languages. A high degree of inconsistency among these responses indicates uncertainty in the model’s understanding, prompting the system to initiate a retrieval process to gather factual information that may clarify or correct the initial response.
The Rowen Consistency Model~(Rowen~CM) extends this idea by assessing the semantic coherence of responses generated by different models, OpenAI~GPT-3.5 and Qwen-Max-0428\footnote{\url{https://qwenlm.github.io/blog/qwen-max-0428/}}, as described in the original paper. By comparing outputs from a primary language model with those generated by a verification model, final consistency model score calculated.
Rowen Hybrid - the hybrid version of Rowen~CL and Rowen-CM, if the sum of the consistency scores for both CL and CM is greater than the threshold, the retriever is used to mitigate potential hallucinations. 

To ensure a reproducible and comparable evaluation of our work, we have reimplemented Rowen approach using LLaMA3.1-8b-instruct as the primary model and Qwen~2.5-72B-Instruct~\cite{qwen2, qwen2.5} as the verification model for consistency model evaluation.

\paragraph{SeaKR}~\cite{DBLP:journals/corr/abs-2406-19215} -- \textbf{Se}lf-\textbf{a}ware \textbf{K}nowledge \textbf{R}etrieval for Adaptive RAG uses an uncertainty approach to minimise hallucinations in LLMs. SeaKR uses the model's internal states to extract self-aware uncertainty, activating external knowledge sources only when the LLM exhibits high uncertainty during generation. This selective retrieval mechanism increases the accuracy and reliability of the generated output.

The SeaKR Uncertainty Module~(SeaKR~UM) is a core component that monitors the internal states of the LLM to quantify its self-aware uncertainty. When the uncertainty level exceeds a predefined threshold, SeaKR~UM triggers the retrieval process to retrieve relevant knowledge snippets from external databases. To ensure the most effective integration of the retrieved information, the SeaKR Re-ranking Component~(SeaKR~RC) re-orders the retrieved snippets based on their ability to reduce the model's uncertainty, selecting the snippet that provides the greatest clarity and factual accuracy.

To ensure a reproducible and comparable evaluation of our approach, we have reimplemented the SeaKR model using Llama-3.1-8b-instruct for the evaluation of self-conscious uncertainty. For consistency, we use the same eigenscore threshold as in the original paper because it gave the best results, but we have also tried others. 


\section{Correlations between evaluation metrics across each dataset}
\begin{table}[th!]
    \centering
    \begin{tabular}{lrrrrrr}
    \toprule
     & InAcc & EM & F1 & Acc & AUC & Corr \\
    \midrule
    InAcc & 1.00 & 0.63 & 0.75 & 0.09 & -0.02 & 0.05 \\
    EM & 0.63 & 1.00 & 0.93 & -0.12 & 0.09 & 0.09 \\
    F1 & 0.75 & 0.93 & 1.00 & -0.06 & 0.08 & 0.09 \\
    Acc & 0.09 & -0.12 & -0.06 & 1.00 & 0.21 & 0.15 \\
    AUC & -0.02 & 0.09 & 0.08 & 0.21 & 1.00 & 0.79 \\
    Corr & 0.05 & 0.09 & 0.09 & 0.15 & 0.79 & 1.00 \\
    \bottomrule
    \end{tabular}
    \caption{Spearman correlations between evaluation metrics normalized across each dataset. The result reveals a low correlation between downstream metrics (InAcc, EM, F1) and self-knowledge metrics (Acc, AUC, Corr). This underscores the importance of conducting a more comprehensive evaluation of self-knowledge of adaptive retrieval systems, rather than relying solely on downstream performance.}
    \label{tab:metric_corr}
\end{table}

\begin{table*}[ht!]
    \centering
    \small
    \resizebox{\textwidth}{!}{

    \begin{tabular}{lccc|ccc|ccc|ccc|ccc|ccc}
    \toprule
    \multirow{2}{*}{{Method}} & \multicolumn{3}{c|}{NQ} & \multicolumn{3}{c|}{SQUAD} & \multicolumn{3}{c|}{TQA} & \multicolumn{3}{c|}{2Wiki} & \multicolumn{3}{c|}{HotPot} & \multicolumn{3}{c}{Musique} \\
    \cmidrule{2-19}
      & InAcc $\uparrow$ & LMC $\downarrow$ & RC $\downarrow$ & InAcc $\uparrow$ & LMC $\downarrow$ & RC $\downarrow$ & InAcc $\uparrow$ & LMC $\downarrow$ & RC $\downarrow$ & InAcc $\uparrow$ & LMC $\downarrow$ & RC $\downarrow$ & InAcc $\uparrow$ & LMC $\downarrow$ & RC $\downarrow$ & InAcc $\uparrow$ & LMC $\downarrow$ & RC $\downarrow$ \\
    \midrule
    Never RAG & 0.446 & 1.0 & 0.00 & 0.176 & 1.0 & 0.00 & 0.636 & 1.0 & 0.00 & 0.318 & 1.0 & 0.00 & 0.286 & 1.0 & 0.00 & 0.106 & 1.0 & 0.00 \\

    Always RAG & 0.496 & 1.0 & 1.00 & 0.312 & 1.0 & 1.00 & 0.610 & 1.0 & 1.00 & 0.374 & 1.0 & 1.00 & 0.410 & 1.0 & 1.00 & 0.100 & 1.0 & 1.00 \\
    
    \midrule
    AdaptiveRAG & \textbf{0.496} & 2.0 & 0.98 & 0.286 & 2.0 & 0.97 & 0.636 & 1.5 & 0.54 & 0.454 & 5.2 & 2.64 & \textbf{0.44} & 4.7 & 2.41 & \textbf{0.154} & 3.6 & 3.63 \\
    DRAGIN & 0.480 & 4.5 & 2.24 & \textbf{0.298} & 4.3 & 2.14 & \textbf{0.667} & 4.0 & 2.0 & \textbf{0.456} & 5.8 & 2.92 & 0.435 & 5.1 & 2.5 & 0.134 & 6.3 & 3.15 \\
    FLARE & 0.462 & 4.26 & 2.0 & 0.288 & 3.2 & 2.5 & 0.648 & 2.1 & 1.39 & 0.424 & 3.9 & 2.85 & 0.372 & 5.1 & 4.07 & 0.106 & 4.3 & 3.11 \\
    Seakr & 0.406 & 14.6 & 1.00 & 0.286 & 14.0 & 1.00 & 0.656 & 14.6 & 1.00 & 0.398 & 12.3 & 2.44 & 0.424 & 9.9 & 1.76 & 0.118 & 12.3 & 2.40 \\

    \midrule
    Ideal & 0.608 & 1.6 & 0.55 & 0.360 & 1.8 & 0.82 & 0.736 & 1.4 & 0.36 & 0.500 & 1.7 & 0.68 & 0.460 & 1.7 & 0.71 & 0.164 & 1.9 & 0.89 \\

    \bottomrule
    \end{tabular}
    }

    \caption{QA Performance of adaptive retrieval fine-tuned with in-domain data. `Ideal' represents the performance of a system with an oracle providing ideal predictions for the need to retrieve. `InAcc' denotes In-Accuracy, measuring the QA system's performance. `LMC' indicates the mean number of LM calls per question, and `RC' represents the mean number of retrieval calls per question. AdaptiveRAG and DRAGIN methods show the best performance.}
    \label{tab:main_ID_results}

\end{table*}

\end{document}